%% file: arxiv.tex
\documentclass{article} 
\pdfoutput=1
\usepackage{a4wide}
\usepackage{color}
\usepackage{hyperref}
\usepackage{url}

\usepackage{ucbpseudocode}
\usepackage{amsmath,amssymb}
\usepackage{amsthm}
\newtheorem{pro}{Proposition}

\newtheorem{thm}{Theorem}

\newtheorem*{pro*}{Proposition}
\newtheorem*{lem*}{Lemma}
\newtheorem*{thm*}{Theorem}
\newtheorem*{cor*}{Corollary}

\newtheorem{innerlabelpro}{Proposition}
\newenvironment{labelpro}[1]
  {\renewcommand\theinnerlabelpro{#1}\innerlabelpro}
  {\endinnerlabelpro}

\newtheorem{innerlabelthm}{Theorem}
\newenvironment{labelthm}[1]
  {\renewcommand\theinnerlabelthm{#1}\innerlabelthm}
  {\endinnerlabelthm}

\usepackage{stmaryrd}
\usepackage{subfigure}
\usepackage{graphicx}
\usepackage{float}
\newfloat{algorithm}{t}{lop}

\newcommand{\fixme}[1]{}
\newcommand{\todo}[1]{}
\newcommand{\checkme}[1]{}

\title{A Greedy Homotopy Method\\ for Regression with Nonconvex Constraints}

\date{\vspace{-1mm}$24^{\footnotesize \mbox{th}}$ of October 2014\vspace{-1mm}}
\author{\begin{tabular}{ccc}
    Fabian L. Wauthier &~~~~~~~~&Peter Donnelly\\
    {\tt flw@well.ox.ac.uk} &&{\tt donnelly@well.ox.ac.uk}
\end{tabular}\\
\\
WTCHG and Department of Statistics\\
University of Oxford 
}

\begin{document}
\maketitle
\begin{abstract}
Constrained least squares regression is an essential tool for high-dimensional data analysis. Given a
partition $\mathcal{G}$ of input variables, this paper considers a particular class of nonconvex constraint
functions that encourage the linear model to select a small number of variables from a small number of groups
in \(\mathcal{G}\). Such constraints are relevant in many practical applications, such as Genome-Wide Association
Studies (GWAS). Motivated by the efficiency of the Lasso homotopy method~\cite{efron04lars,osborne00homotopy},
we present RepLasso, a greedy homotopy algorithm that tries to solve the induced sequence of nonconvex
problems by solving a sequence of suitably {\it adapted} convex surrogate problems. We prove that in some
situations RepLasso recovers the global minima of the nonconvex problem. Moreover, even if it does not recover
global minima, we prove that in relevant cases it will still do no worse than the Lasso in terms of support and signed
support recovery, while in practice outperforming it. We show empirically that the strategy can also be used
to improve over other Lasso-style algorithms. Finally, a GWAS of ankylosing spondylitis highlights our
method's practical utility.
\end{abstract}

\input{introduction.tex}

\input{related.tex}
\input{algorithm.tex}

\input{analysis.tex}
\input{results.tex}

\input{conclusion.tex}
\input{acknowledgements.tex}
\newpage
\bibliography{bibliography}
\newpage
\appendix

\input{supplementary.tex}
\end{document}

%% file: introduction.tex
\section{Introduction}\label{sec:introduction}
We are interested in model parsimony in the context of linear observation
models of the form
\begin{align}
y = X\beta^* + w~~~~~~w \sim \mathcal{N}(0, \sigma^2 I),\label{eq:genmodel}
\end{align}
where \(X\) is an \(n\times p\) matrix of covariates, \(\beta^*\) is a regression parameter, \(w\) is a noise
vector and \(y\) is a vector of responses. Given \(X, y\), a constraint function \(\Omega(\cdot)\), and
constraint parameter \(\tau > 0\), constrained least squares regression estimates \(\beta^*\) as
\begin{align}
\underset{\beta\in \mathbb{R}^p}{\mbox{argmin}} \frac{1}{2n}\left|\!\left|y -
  X\beta\right|\!\right|_2^2~~~ \mbox{s.t.}~~~\Omega(\beta) \leq \tau.\tag{$G$}\label{eq:general}
\end{align}
A closely related formulation writes~(\ref{eq:general}) in penalized form using penalizer \(\Omega(\cdot)\)
and penalty parameter \(\lambda > 0\). In the following, we will motivate our algorithm using the constrained
formulation~(\ref{eq:general}). However, the proposed algorithm is then more naturally expressed as solving a sequence of
penalized problems. 

The use of sparsity for model selection is an integral component of the modern statistics toolbox and is
especially relevant for the \(n \ll p\) case. The Lasso~\cite{tibshirani94lasso} is a well-known special case
of~(\ref{eq:general}) which replaces a hard \(\ell_0\) constraint (i.e., \(|\!|\beta|\!|_0 \leq \tau\)) by an
\(\ell_1\) surrogate (i.e., \(\left|\!\left|\beta\right|\!\right|_1 \leq \tau\)) that retains some
sparsity-inducing properties. Since the Lasso regularization path is continuous and piecewise
linear~\cite{rosset2007piecewise}, it can be easily traced out using the homotopy
method~\cite{efron04lars,osborne00homotopy}. This is done by writing~(\ref{eq:general}) in penalized form and
then tracing out \(\lambda = \infty \downarrow 0\). In the following, we will often refer to the Lasso
homotopy method simply as the Lasso. The efficiency of the homotopy method is one of the main benefits
of the \(\ell_1\) relaxation approach and is key for efficient model selection.


Recently, there has been increased interest in enhancing the Lasso with structured sparsity. Relevant examples
include~\cite{jenatton:inria-00377732,kim12tree,kowalski2009sparsity,yuan2006model,zhou2010exclusive}. In all
these cases the structured sparsity is induced by replacing the \(\ell_1\) penalizer of the Lasso with more
complex, yet still convex, penalizers. Because the overall objective remains convex in \(\beta\), efficient
algorithms exist to solve these problems. While these methods have many practical applications, the focus on
convex formulations has necessarily excluded important inference problems that cannot easily be phrased in
terms of structured convex objectives. This paper was motivated by applications where for a given partition
\(\mathcal{G}\) of variables \(\{1, \ldots, p\}\) it is reasonable to assume that in {\it at most} a few
groups {\it at most} a few {\it representatives} are relevant for predicting \(y\). For instance, in the Genome-Wide
Association Study (GWAS) we consider in Section~\ref{sec:results}, it is reasonable to suppose that in at most
a small number of genes at most a small number of SNPs are associated with the response variable. To suite
these and other applications, we are interested in constraint functions \(\Omega(\cdot)\) that encourage the
estimate of \(\beta^*\) to be nonzero on at most a few elements in at most a few groups \(G \in \mathcal{G}\),
which can be thought of as orthogonal to the Group Lasso~\cite{yuan2006model}. This problem is not adequately
solved by the Exclusive/Elitist Lasso~\cite{kowalski2009sparsity,zhou2010exclusive}, which is a convex
formulation that generally selects {\it at least one} but at most a few variables from each group. This is
problematic if we believe that most groups in \(\mathcal{G}\) will contain no relevant variables for
predicting \(y\), as for example in the GWAS application. To the best of our knowledge the selection
behavior we seek can only be achieved by nonconvex constraints.

Given a parameter \(\theta \in \mathbb{R}^p\) and the partition \(\mathcal{G}\), we will encode our
constraints as a nonconvex function \(\Omega_{\theta,\mathcal{G}}(\cdot)\).  For fixed penalty parameter
\(\lambda > 0\), there are specialized methods for the nonconvex penalized cousin of~(\ref{eq:general})
(e.g.,~\cite{candes2008enhancing,fan01scad,hunter2005variable,loh13mest,zou2008one}). While these methods
might be appropriate for finding a local minimum of~(\ref{eq:general}) with
\(\Omega_{\theta,\mathcal{G}}(\cdot)\) and some \(\tau\) fixed, they are not useful for developing a
homotopy-like algorithm which allows \(\tau\) to range over an interval. Motivated by the practicality and
efficiency of the Lasso homotopy method, we propose RepLasso (for ``Representative Lasso''), a homotopy-like
algorithm that attempts to fill this gap. At a high level, RepLasso tries to build and solve a sequence of
convex surrogates so that, as \(\tau\) is swept out, the boundary of the surrogate constraint ball locally
approximates the boundary of the ball induced by \(\Omega_{\theta,\mathcal{G}}(\cdot)\). A crucial feature
that allows us to do this efficiently is that the nonconvex constraint balls induced by
\(\Omega_{\theta,\mathcal{G}}(\cdot)\) can be decomposed as unions of convex balls. Moreover, the sequence of
surrogates is chosen so that the induced regularization path is continuous and piecewise linear and can thus
be efficiently traced out using a homotopy-style algorithm.

To motivate the algorithm, we show theoretically that, under certain conditions, RepLasso traces out the global
minima of~(\ref{eq:general}) with constraint function \(\Omega_{\theta,\mathcal{G}}(\cdot)\). More importantly, we
prove that, even though RepLasso may not exactly solve this problem in general, on relevant problems it will still do at
least as well as the Lasso in terms of support subset and signed support recovery. In practice, a strict
improvement is observed. A class of Lasso-style algorithms has recently been popularized which pre-process
\(X, y\) in some way, prior to solving a standard Lasso problem
(e.g.,~\cite{huang11corrsift,jia12whitened,paul08plasso,zou2006adaptive}). As we demonstrate in
Section~\ref{sec:results}, RepLasso can also yield strict improvements in these settings. Furthermore,
RepLasso can be usefully applied to \(\ell_1\) constrained logistic regression~\cite{lee2006efficient}, as we
demonstrate in a GWAS application. Lastly, we prove in the Supplementary Material that, given some mild
assumptions, a variant of RepLasso cannot do worse than the well-known Lars algorithm of Efron et
al.~\cite{efron04lars}.

The paper is organized as follows: We review related research in Section~\ref{sec:related} before introducing
\(\Omega_{\theta,\mathcal{G}}(\cdot)\) and simplifying~(\ref{eq:general}) in Section~\ref{sec:problem}. In
Section~\ref{sec:replasso} we present the RepLasso as a generalization of the Lasso homotopy method and in
Section~\ref{sec:analysis} give a theoretical comparison of Lasso and RepLasso. Results on synthetic data and
a GWAS application are given in Section~\ref{sec:results}. We conclude with final remarks in
Section~\ref{sec:conclusion}. Proofs are collected in the Supplementary Material.

%% file: related.tex
\section{Related Research}\label{sec:related}
Nonconvex penalties for least squares regression have been (for example) considered by Fan and
Li~\cite{fan01scad} and Zhang~\cite{zhang2010nearly}. Methods for optimizing convex loss functions with
nonconvex regularizers include, among others, local quadratic approximation~\cite{fan01scad},
minorization-maximization~\cite{hunter2005variable}, local linear approximation~\cite{zou2008one} and
composite gradient descent~\cite{loh13mest}. The Adaptive Lasso of~\cite{candes2008enhancing} is also related
to our method.  However, a drawback of all of these approaches is that they focus on a single optimization
problem, indexed by a fixed penalty parameter. This precludes their use for efficiently minimizing a sequence
of problems~(\ref{eq:general}) indexed by \(\tau\), as in a homotopy method. The various applications of the
homotopy idea have so far focused on other convex problems. Well-known examples are the Elastic
Net~\cite{zou05regularization} and the SVM~\cite{hastie2004entire}. However, there are very few extensions to
nonconvex least squares problems. One of the few methods that efficiently sweeps out local minima paths of
such problems is due to Zhang~\cite{zhang2010nearly}. However, as~\cite{zhang2010nearly} assumes the penalty
to be separable across the \(p\) coefficients, it is not useful for the type of structured sparsity we
consider in this paper. There has been growing interest in more complex sparsity patterns induced by
structured penalties. Among convex extensions, the Group Lasso~\cite{yuan2006model} is a well-known example,
which for some partition \(\mathcal{G}\), replaces the \(\ell_1\) penalty above by a sum of \(\ell_2\)
penalties over groups of variables indexed by \(G \in \mathcal{G}\). This method will select groups of
variables, not representatives, and so can be seen as a counterpart to the work in this paper. Several
variations of this approach have been proposed~\cite{jenatton:inria-00377732,kim12tree}. The Exclusive/Elitist
Lasso~\cite{kowalski2009sparsity,zhou2010exclusive} is more closely aligned with our goal. However, this
method effectively encourages each group to contribute at least one variable to the support set. In contrast,
our method encourages the selection of a small number of variables in a small number of groups. Finally, while
there are other structured, nonconvex penalizers (e.g.,~\cite{wang2007group}), there are no homotopy
algorithms to solve them.

%% file: algorithm.tex
\section{Structured Nonconvex Problems}\label{sec:problem}
Many situations exist where for some partition \(\mathcal{G}\) of \(\{1, \ldots, p\}\) we know
that \(\beta^*\) contains at most a few nonzero elements in at most a few groups \(G \in
\mathcal{G}\).\footnote{In the GWAS application in Section~\ref{sec:results}, \(\mathcal{G}\) corresponds
  to a partition of SNPs by genes and we know that in at most a few genes at most a few SNPs are truly
  relevant for predicting \(y\).} Given a partition \(\mathcal{G} = \{G_1, \ldots, G_g\}\) without singleton
or empty sets, and a vector \(\theta = (\theta_1, \ldots, \theta_g) \geq 0\), the following constraint
function targets this situation
\begin{align}
\Omega_{\theta,\mathcal{G}}(\beta)\!&=\!\!\!\!\!\sum_{i<j \in G_{g'} \in \mathcal{G}}\!\!\!\frac{\omega_{\theta_{g'}}(\beta_i, \beta_j)}{|G_{g'}| -
  1}\label{eq:starshapedpen}\\
\omega_{\theta_{g'}}(\beta_i, \beta_j) &= \min(|\beta_i|,|\beta_j|)(1 + \theta_{g'}) + \max(|\beta_i|,|\beta_j|).\nonumber
\end{align}
\todo{diff submodular fns} Let \(B_{\theta, \mathcal{G}}(\tau) = \left\{\beta \in \mathbb{R}^p:
\Omega_{\theta, \mathcal{G}}(\beta) \leq \tau\right\}\) be the induced constraint balls. We are interested in
the following nonconvex instance of~(\ref{eq:general}) with the constrained objective \(J_\tau(\beta)\) over
\(\beta\), indexed by \(\tau\)
\begin{align}
\beta(\tau) &\in 
\underset{\beta\in \mathbb{R}^p}{\mbox{argmin}} J_\tau(\beta) = \underset{\beta\in \mathbb{R}^p}{\mbox{argmin}} \left\{\begin{array}{cl}
\frac{1}{2n}\left|\!\left|y - X\beta\right|\!\right|_2^2 &\mbox{if}~~\beta \in B_{\theta, \mathcal{G}}(\tau)\\
\infty & \mbox{o.w.} 
\end{array}
\right.\!\!\!\!.\tag{$P1$}\label{eq:nonconvex}
\end{align}
\begin{figure*}[t]
\begin{center}
\subfigure[]{
\includegraphics[width=6cm]{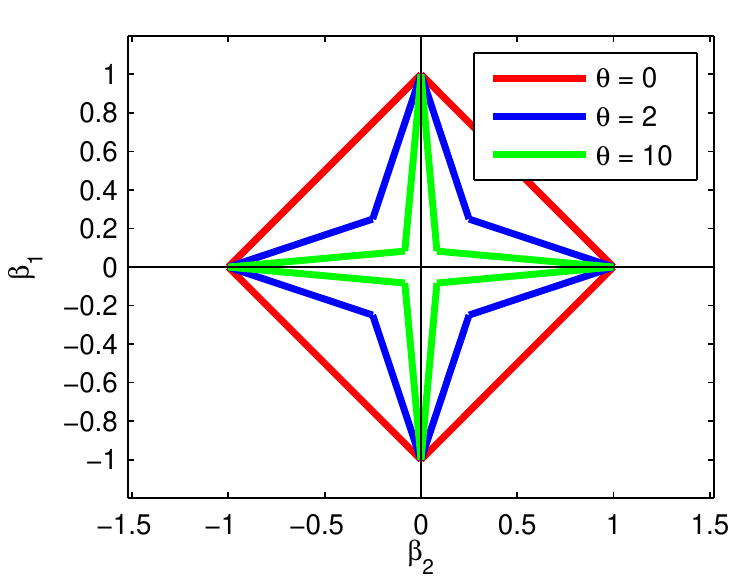}\label{subfig:starshapedset2d}
}
\subfigure[]{
\includegraphics[width=6cm]{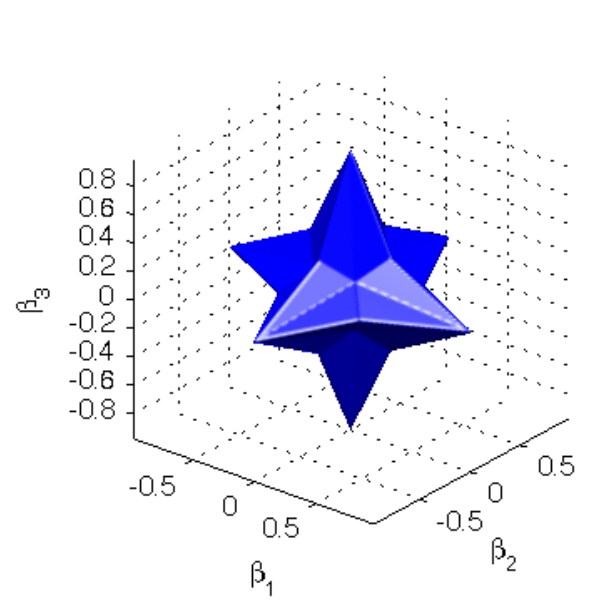}\label{subfig:starshapedset3d}
}
\caption{~\subref{subfig:starshapedset2d} The balls induced by
  Eq.~(\ref{eq:starshapedpen}) for \(\mathcal{G} = \{\{1,2\}\}\) and varying values of
  \(\theta\). If \(\theta = 0\)
  we recover the \(\ell_1\) norm as a special case.~\subref{subfig:starshapedset3d} A ball
  for \(\mathcal{G} = \{\{1,2,3\}\}, \theta = 2\).}\label{fig:stars}
\end{center}
\end{figure*}
If \(\theta = {\bf 0}\), then \(\Omega_{\theta,\mathcal{G}}(\beta) = |\!|\beta|\!|_1\) for all
\(\mathcal{G}\), and so~(\ref{eq:nonconvex}) recovers the Lasso problem as special case. However, when
\(\theta \neq {\bf 0}\), the constraint function is non-separable, non-convex and induces star-shaped
balls, as exemplified in Figure~\ref{fig:stars}. Note that
when a subset of components of \(\theta\) is set to zero, we can effectively treat the variables corresponding
to those groups as ungrouped, as they only contribute an \(\ell_1\) penalty to
\(\Omega_{\theta,\mathcal{G}}(\beta)\).  When \(\theta \neq {\bf 0}\),
\(\Omega_{\theta,\mathcal{G}}(\beta)\) can be thought of as a nonconvex counterpart to the well-known Group
Lasso penalty~\cite{yuan2006model}, where the nonconvexity encourages solutions \(\beta(\tau)\)
of~(\ref{eq:nonconvex}) to select {\it at most} a small number of {\it representatives} from {\it at most} a
few groups \(G \in \mathcal{G}\). The penalty \(\Omega_{\theta,\mathcal{G}}(\beta)\) is also distinct from the
convex Exclusive/Elitist Lasso penalty~\cite{kowalski2009sparsity,zhou2010exclusive}, which effectively
encourages each group to select at least one variable. 

For some positive vector \(s\), let \(B_s(\tau) = \{\beta \in \mathbb{R}^p: |\!|\mbox{diag}(s)\beta|\!|_1 \leq
\tau\}\) be the \(s\)-weighted \(\ell_1\) ball. An important property that is suggested by
Figure~\ref{fig:stars} is that \(B_{\theta,\mathcal{G}}(\tau)\) can be written as a union of weighted
\(\ell_1\) balls and so has planar faces. Let \(\Gamma(i) \in \{1, \ldots, g\}\) be the (unique) variable
index so that \(i \in G_{\Gamma(i)}\).

\begin{figure*}[t]
\begin{center}
\subfigure[]{
\includegraphics[width=5.6cm]{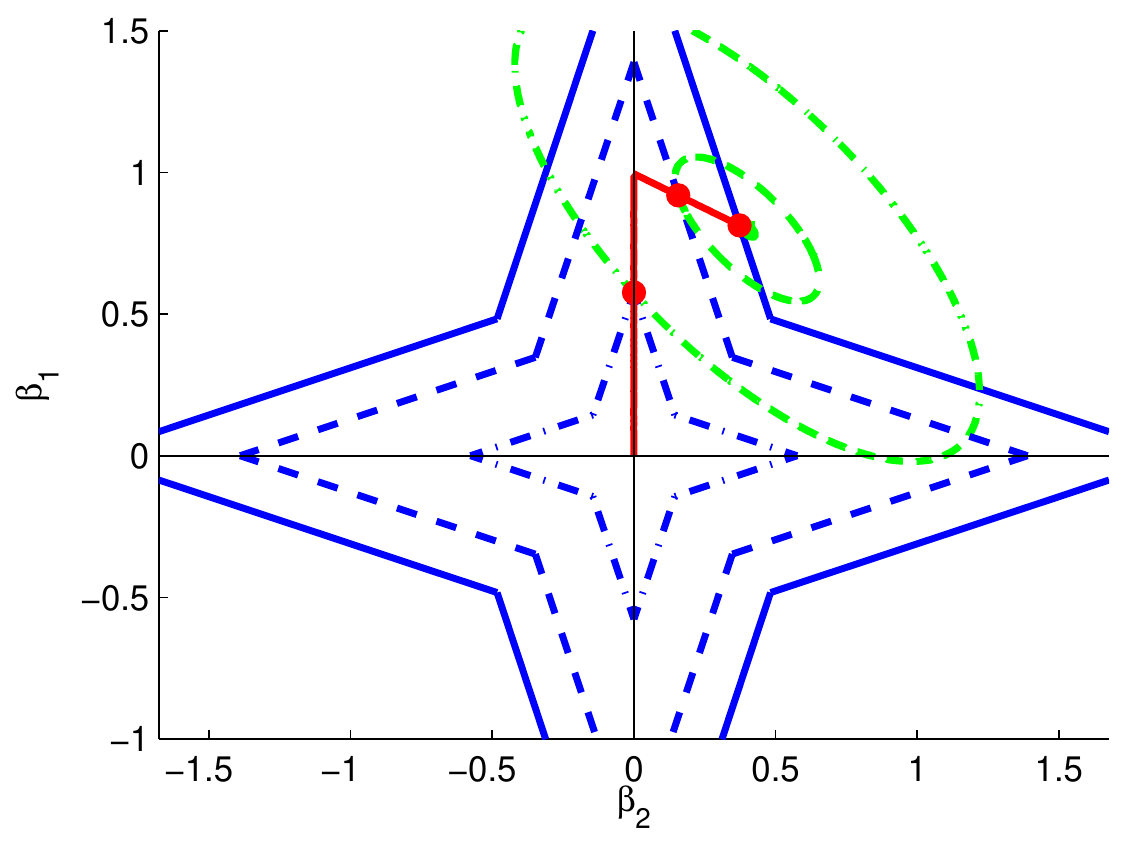}\label{subfig:starshapedpath}
}
\subfigure[]{
\includegraphics[width=5.6cm]{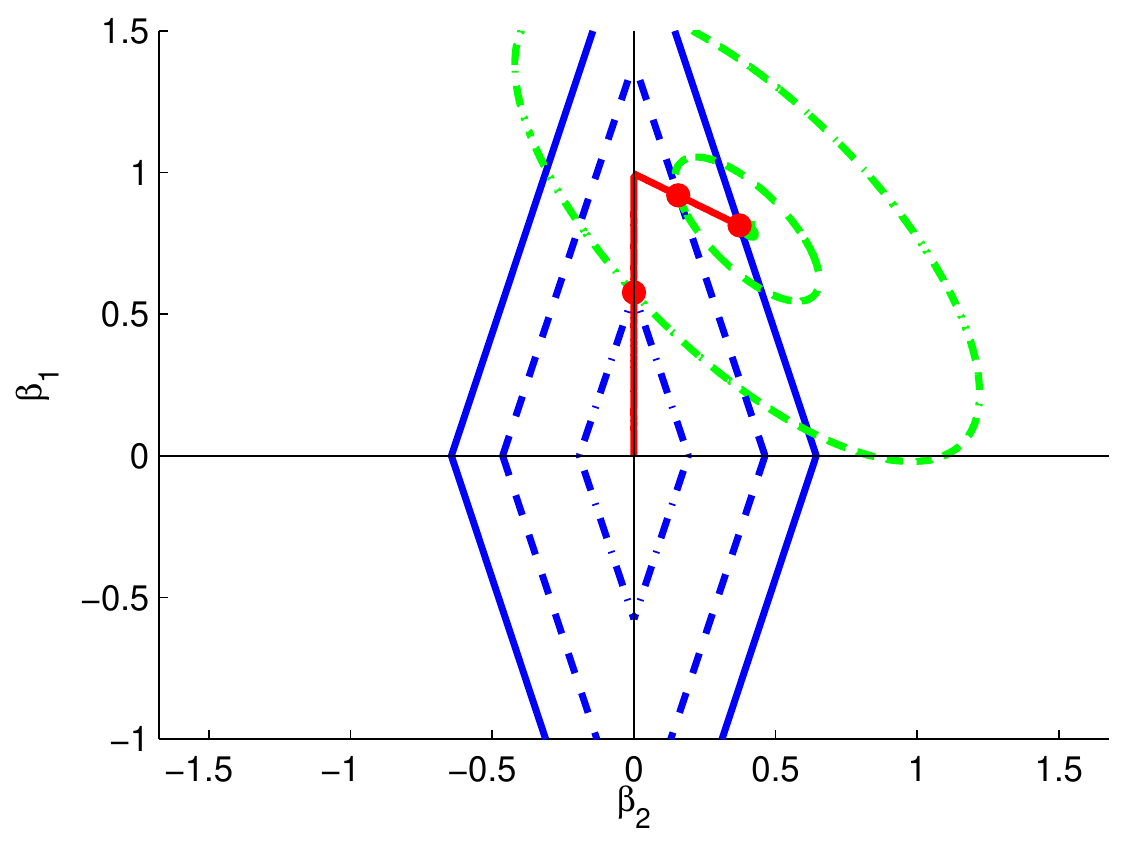}\label{subfig:diamondpath}
}
\end{center}
\caption{~\subref{subfig:starshapedpath} The global minima path for \(B_{2,\{\{1,2\}\}}(\tau)\). The
  dashed-dotted, dashed and solid constraint boundaries correspond to progressively larger \(\tau\). The
  solutions on the regularization path (red) are defined by intersections of the constraint set with the
  corresponding ellipsoid (red dots).~\subref{subfig:diamondpath} The corresponding regularization path for
  the weighted \(\ell_1\) ball \(B_{(1, 3)^\top}(\tau)\). If the weighting is known the same regularization
  path can be reproduced.}
\label{fig:paths}
\end{figure*}

\begin{pro}[Union Decomposition]\label{pro:decomposition}
Let the partition be \(\mathcal{G} = \{G_1, \ldots, G_g\}\) and the parameter \(\theta = (\theta_1, \ldots,
\theta_g) \geq 0\). There is a finite set \(\mathcal{S}_{\theta,\mathcal{G}} \subset \mathbb{R}^p\) of vectors
\(s \geq {\bf 1}\), so that for any \(\tau > 0\)
\begin{align}
B_{\theta, \mathcal{G}}(\tau) = \bigcup_{s \in \mathcal{S}_{\theta,\mathcal{G}}} B_{s}(\tau).
\end{align}
Define \(\Pi_{g'}\) to be all permutations \(\pi_{g'}\) of the elements in \(G_{g'}\) and let \(\Pi_{\mathcal{G}} =
\times_{g' = 1}^g \Pi_{g'}\) be their cross-product, whose elements \(\pi \in \Pi_{\mathcal{G}}\) are \(g\)-tuples of
permutations \(\pi = (\pi_1, \ldots, \pi_g)\). For some \(\pi \in \Pi_{\mathcal{G}}\), denote by \(\pi_{\Gamma(i)}(i) \in
\{1 , \ldots, |G_{\Gamma(i)}|\}\) the position of \(i \in G_{\Gamma(i)}\) in permutation \(\pi_{\Gamma(i)}\).  We have
\begin{align}
\mathcal{S}_{\theta,\mathcal{G}} &= \cup_{\pi \in \Pi_{\mathcal{G}}} \{s_\pi\}\\
s_{\pi, i} &= 1 + (\pi_{\Gamma(i)}(i)-1) \frac{\theta_{\Gamma(i)}}{|G_{\Gamma(i)}| - 1}.
\end{align}
\end{pro}
A common, brute force approach that eliminates the computational issues of~(\ref{eq:nonconvex}) replaces
\(B_{\theta, \mathcal{G}}(\tau)\) by its convex hull. However, in this case the convex hull is the \(\ell_1\)
ball \(B_{1}(\tau)\) which eliminates all structural information inherited from \(\theta, \mathcal{G}\) and
would not lead to the desired selection behavior. In this paper we advocate an orthogonal strategy that instead focuses on replacing
\(B_{\theta, \mathcal{G}}(\tau)\) by a suitable {\it sequence} of {\it weighted} \(\ell_1\) balls, indexed by
\(\tau\). To achieve this, we will exploit the decompositional structure of \(B_{\theta, \mathcal{G}}(\tau)\)
highlighted in Proposition~\ref{pro:decomposition}. Specifically, our method is motivated by the following
extension of a well-known result of Rosset and Zhu~\cite{rosset2007piecewise}.
\begin{pro}[Local Piecewise Linearity]\label{pro:locallinear}
Suppose \(X\) has absolutely continuous distribution and that \(\exists \tau' > 0\)
s.t. \(\nexists \beta \in B_{\theta, \mathcal{G}}(\tau')\) which is a minimum of \(|\!|y -
X\beta|\!|_2^2\). Let \(\tau_{\max}\) be the supremum over these \(\tau'\). The set of local minima of \(J_\tau(\cdot)\) in~(\ref{eq:nonconvex}) with \(\tau \in (0,
\tau_{\max})\) is with probability 1 a finite union of piecewise linear paths, each path indexed by \(\tau\)
  and lying on a ball \(B_{s}(\tau), s \in \mathcal{S}_{\theta,\mathcal{G}}\). 
\end{pro}

Proposition~\ref{pro:locallinear} emphasizes that for the range of interesting values of \(\tau
\in (0, \tau_{\max})\), the local minima of \(J_\tau(\cdot)\) in~(\ref{eq:nonconvex}) can be grouped into a
set of local minima {\it paths}, each indexed by \(\tau\). Moreover, any such local minimum path lies on some
weighted \(\ell_1\) ball \(B_{s}(\tau)\), with \(s \in \mathcal{S}_{\theta,\mathcal{G}}\) appropriately
chosen. With the aid of Proposition~\ref{pro:locallinear}, it is possible to re-express~(\ref{eq:nonconvex})
as a special set of penalized optimization problems, indexed by \(\tau\). This change of representation will
be useful for the homotopy-like algorithm we present shortly.  By Proposition~\ref{pro:locallinear} and
convexity, for any solution \(\beta(\tau)\) of~(\ref{eq:nonconvex}) with \(\tau \in (0, \tau_{\max})\) (i.e.,
a global minimum of \(J_\tau(\cdot)\)), \(\exists \lambda^*(\tau), s^*(\tau) \in
\mathcal{S}_{\theta,\mathcal{G}}\) so that
\begin{align}
\!\!\!\!\!\beta(\tau)\!&\in\!\underset{\beta\in \mathbb{R}^p}{\mbox{argmin}} \frac{1}{2n}\!\left|\!\left|y\!-\!X\beta\right|\!\right|_2^2\!+\!\lambda^*\!(\tau)\!\left|\!\left|\mbox{diag}(\!s^*\!(\tau)) \beta\right|\!\right|_1\!.\!\!\tag{$P2$}\label{eq:nonconvex_equiv}
\end{align}
Thus, modulo uniqueness issues, there
exist \(\lambda^*(\tau)\), \(s^*(\tau)\) so that~(\ref{eq:nonconvex_equiv}) is in some sense equivalent
to~(\ref{eq:nonconvex}). Figure~\ref{fig:paths} shows a motivating example of this. In this case, the global
minimum path of Figure~\ref{subfig:starshapedpath} could be reproduced using the \(B_{(1, 3)^\top}(\tau)\)
balls of Figure~\ref{subfig:diamondpath}. Of course, knowledge of the vector-valued function \(s^*(\tau) \in \mathcal{S}_{\theta,\mathcal{G}}\) would imply knowing for each \(\tau\)  roughly where on \(B_{\theta,
  \mathcal{G}}(\tau)\) the global minimum of \(J_\tau(\cdot)\) in~(\ref{eq:nonconvex}) lies, which is hard in
general. We thus cannot expect to be able to efficiently produce the {\it entire} regularization path
of~(\ref{eq:nonconvex}) for all \(\tau \in (0, \tau_{\max})\) using the equivalence
between~(\ref{eq:nonconvex_equiv}) and~(\ref{eq:nonconvex}).

\paragraph{A Simplifying Assumption.} 
The formulation in~(\ref{eq:nonconvex_equiv}) replicates the regularizing effect of \(B_{\theta,
  \mathcal{G}}(\tau)\) in~(\ref{eq:nonconvex}) using a sequence of weighted \(\ell_1\) balls that depend on
\(\tau\) (characterized by \(s^*(\tau)\)). This dependence is necessary as the global minimum of
\(J_{\tau}(\cdot)\) in~(\ref{eq:nonconvex}) can ``jump'' from one weighted \(\ell_1\) ball to another as we
vary \(\tau \in (0, \tau_{\max})\). If we let \(S(\beta)\) be the support of some vector \(\beta\), we can simplify the
problem of finding sequences \(\lambda^*(\tau), s^*(\tau)\) for~(\ref{eq:nonconvex_equiv}), by assuming that
\begin{itemize}
\item[{\bf A0}:] \(\exists s^* \in \mathcal{S}_{\theta,\mathcal{G}}\) so that \(\forall \tau \in (0, \tau_{\max})\), 
 (\ref{eq:nonconvex}) has a unique solution which lies in \(B_{s^*}(\tau)\). For any \(0 < \tau_1 <
  \tau_2 < \tau_{\max}\), the solutions of~(\ref{eq:nonconvex}) satisfy \(S(\beta(\tau_1)) \subseteq
  S(\beta(\tau_2))\).
\end{itemize} 
Under {\bf A0}, the problem immediately reduces to finding a sequence \(\lambda^*(\tau)\) and a single
positive vector \(s^* \in \mathcal{S}_{\theta,\mathcal{G}}\). In fact, it is not even necessary to know the
precise function \(\lambda^*(\tau)\): For any \(\lambda > 0\), so long as the solution \(\bar \beta(\lambda)\)
to~(\ref{eq:nonconvex_equiv}) with \(\lambda^*(\tau)\) replaced by \(\lambda\) and \(s^*(\tau) = s^*\),
satisfies for \(\tau \triangleq \left|\!\left|\mbox{diag}(s^*) \bar \beta(\lambda)\right|\!\right|_1\) that \(\tau
\in (0, \tau_{\max})\), we know that \(\lambda = \lambda^*(\tau)\).  Thus, under {\bf A0} we only seek to find
the vector \(s^* \in \mathcal{S}_{\theta,\mathcal{G}}\) so that solving~(\ref{eq:nonconvex}) is for some
\(\lambda\) equivalent to solving
\begin{align}
\bar \beta(\lambda) &\in \underset{\beta\in \mathbb{R}^p}{\mbox{argmin}} \frac{1}{2n}\left|\!\left|y-X\beta\right|\!\right|_2^2+\lambda
\left|\!\left|\mbox{diag}(s^*)\beta\right|\!\right|_1.\tag{$S$}\label{eq:surrogate}
\end{align}
This paper makes two main contributions. The first contribution in Section~\ref{sec:replasso} proves that if
{\bf A0} holds, then there is an algorithm, RepLasso, which (effectively) greedily estimates the vector
\(s^*\) making~(\ref{eq:surrogate}) and~(\ref{eq:nonconvex}) equivalent while sweeping out \(\lambda > 0\) and
producing solutions \(\bar \beta(\lambda)\) in a homotopy-like fashion. Of course, if {\bf A0} does not hold,
there may not be an equivalence between~(\ref{eq:surrogate}) and~(\ref{eq:nonconvex}). In that case, we may think
of~(\ref{eq:surrogate}) as a convex surrogate for~(\ref{eq:nonconvex}) for some vector \(s^*\) that is
greedily constructed by RepLasso. The second contribution of this paper is to prove in
Section~\ref{sec:analysis} that, whether {\bf A0} holds or not, RepLasso will in relevant regression problems
still perform at least as well as the Lasso in terms of variable selection. Empirical evidence in
Section~\ref{sec:results} shows a strict improvement in practice.

\begin{algorithm*}
{\footnotesize
\begin{pseudocode}{REPLASSO}{\(X, y, \mathcal{G}, \theta\)}\label{alg:replasso}
\bar y = 0, A = (), L = 0, \lambda = |\!|X^\top y|\!|_\infty, s(\lambda) = {\bf 1}, \bar\beta(\lambda) =
0\todo{change \(s(\lambda)\) to \(\bar s(\lambda)?, \(A_A\) to \(\bar A_A\), etc}\\
\WHILE \lambda > 0\DO 
\BEGIN
\mbox{Stage 1}\left\{\BEGIN
\IF L = 0~~\mbox{\# Add a variable}\DO
\BEGIN
A = (A, i^*),~\mbox{where}~i^* = \mbox{argmax}_{j \in A^c} \left|X_j^\top (y - \bar y)/s_j(\lambda)\right|\\

s_M(\lambda^-) = s_M(\lambda) + \frac{\theta_{\Gamma(i^*)}}{|G_{\Gamma(i^*)}| - 1}{\bf 1},~~~s_{M^c}(\lambda^-) = s_{M^c}(\lambda),~~\mbox{with}~~M = \left\{A^c \cap G_{\Gamma(i^*)}\right\}\\
\END\\
\IF L = 1~~\mbox{\# Delete a variable}\DO
\BEGIN
A = A \backslash i^*,~\mbox{where}~i^* = \mbox{arg}_{i \in A} \llbracket\bar\beta_{i}(\lambda) = 0\rrbracket\\
\END\\
\END\right.\\
\mbox{Stage 2}\left\{\BEGIN
\bar w_A =  A_A \left( X_A^\top X_A \right)^{-1} \mbox{diag}(\mbox{sgn}\left(X_A^\top (y - \bar y)\right))
s_A(\lambda),~\mbox{with}~A_A~\mbox{s.t.}~|\!|X_A \bar w_A|\!|_2^2 = 1\\
\END\right.\\
\mbox{Stage 3}\left\{\BEGIN
\mbox{Find smallest \(\rho > 0\) s.t.}\\
~~~\bullet~\exists j \in A^c~\mbox{s.t.}~|X_j^\top (y - \bar y - \rho X_A \bar w_A) / s_j(\lambda)| = \lambda -
\rho\!:~\mbox{set}~L =0\\
~~~\bullet~\exists i \in A~\mbox{s.t.}~\bar\beta_i(\lambda) \neq 0~\mbox{and}~\bar\beta_i(\lambda) + \rho w_i=0\!:~\mbox{set}~L = 1\\

\END\right.\\
\mbox{Stage 4}\left\{\BEGIN
\bar\beta_A(\lambda - \rho) = \bar\beta_A(\lambda) + \rho \bar w_A,~~~\bar\beta_{A^c}(\lambda - \rho) = 0,~~~\bar y = X \bar\beta(\lambda - \rho) \\
\lambda = \lambda - \rho\\
\END\right.\\
\END\\
\RETURN \bar\beta
\end{pseudocode}
}
\end{algorithm*}

\section{RepLasso: A Greedy Homotopy Method}\label{sec:replasso}
To motivate our description of RepLasso, we first make the following observation regarding the sensitivity of
problems in the form of~(\ref{eq:surrogate}) to approximations of \(s^*\).  For a positive vector \(b\), let
\(\bar \beta_{b}(\lambda)\) be a solution to~(\ref{eq:surrogate}) with penalty \(\lambda |\!|\mbox{diag}(b)
\beta|\!|_1\).
\begin{pro}[Recoverability of~(\ref{eq:surrogate})]\label{pro:underest}
Suppose \(X\) has absolutely continuous distribution.  For any vectors \(a \geq b \geq {\bf 1}\) and \(\lambda >
0\), with probability 1
\(\bar\beta_{a}(\lambda), \bar\beta_{b}(\lambda)\) are unique. If additionally \(|\!|\mbox{diag}(a)\bar
\beta_{b}(\lambda)|\!|_1 = |\!|\mbox{diag}(b)\bar \beta_{b}(\lambda)|\!|_1\), then \(\bar
\beta_{a}(\lambda) = \bar \beta_{b}(\lambda)\).
\end{pro}
Thus, if \(\bar \beta_{{b}}(\lambda)\) has zero coefficients, then it doesn't matter if on those coefficients
\({b}\) underestimates the value of \(a\), so long as \({b}\) matches \(a\) on the remaining
coefficients. 

The RepLasso algorithm (Algorithm~\ref{alg:replasso}) is a generalization of the Lasso homotopy
method~\cite{efron04lars,osborne00homotopy} that exploits Proposition~\ref{pro:underest} to
solve~(\ref{eq:surrogate}). If \(X\) is absolutely continuous and {\bf A0} holds, then
Proposition~\ref{pro:underest} suggests the existence of a sequence \({s}(\lambda)\), satisfying \(\forall
\lambda > 0, s^* \geq {s}(\lambda) \geq {\bf 1}\), so that with probability 1~(\ref{eq:surrogate}) can
\(\forall \lambda > 0\) be solved as \(\bar \beta(\lambda) \triangleq \bar \beta_{s^*}(\lambda) = \bar
\beta_{{s}(\lambda)}(\lambda)\). As Theorem~\ref{thm:replars_optimality} shows, RepLasso computes such a
sequence \({s}(\lambda)\), while simultaneously producing solutions \(\bar
\beta_{{s}(\lambda)}(\lambda)\). Notice that RepLasso is identical to the Lasso homotopy method if \(\theta =
     {\bf 0}\) (which means that \(\forall \lambda > 0, s(\lambda) = {\bf 1}\)). The only differences are
that \(s(\lambda) \neq {\bf 1}\) when \(\theta \neq {\bf 0}\). We will discuss RepLasso as constructive
proof for Theorem~\ref{thm:replars_optimality}.
\begin{thm}[RepLasso]\label{thm:replars_optimality}
Assume that \(X\) has absolutely continuous distribution and that {\bf A0} holds. Let \(s^* \in
\mathcal{S}_{\theta,\mathcal{G}}\) be the vector so that~(\ref{eq:nonconvex}) is equivalent
to~(\ref{eq:surrogate}). Then with probability \(1\), RepLasso produces a sequence \({s}(\lambda)\) so that
\(\bar \beta_{s^*}(\lambda) = \bar \beta_{{s}(\lambda)}(\lambda)\). By the equivalence
of~(\ref{eq:nonconvex}) and~(\ref{eq:surrogate}), it follows that with probability 1, RepLasso produces the
global minima of~(\ref{eq:nonconvex}).
\end{thm}
\begin{proof}
Note from our earlier discussion that it is sufficient for RepLasso to estimate sequences \({s}(\lambda)\)
which are {\it piecewise constant} with changepoints at values \(\lambda_t\) where the support of \(\bar
\beta_{s^*}(\lambda_t)\) changes. By {\bf A0}, we know that the support of \(\bar \beta_{s^*}(\lambda)\) is
monotonically increasing with \(\lambda\) decreasing. Hence, we only need to discuss the variable addition
case (case \(L = 0\) in stage 1) of RepLasso for this argument.  Conceptually, RepLasso first initializes
\({s}(\infty) = {\bf 1}\) (for practical reasons it suffices to start at \(\lambda = |\!|y^\top
X|\!|_\infty\)). Then, while keeping \(s(\lambda) = s(\infty)\) constant, RepLasso (conceptually) traces out
\(\lambda = \infty \downarrow 0\) while solving \(\bar \beta_{{s}(\lambda)}(\lambda) = {\bf 0}\) until reaching
\(\lambda_1 = |\!|y^\top X|\!|_\infty\), where the first variable \(i_1^*\) is selected by \(\bar
\beta_{{s}(\lambda)}(\lambda)\) (the \(L = 0\) case in stage 1 of RepLasso). Because \(s(\lambda) = {\bf 1}\) was up
to now fixed, RepLasso is up to this point identical to the Lasso homotopy method.  Due to
Proposition~\ref{pro:underest}, we know that with probability 1, \(\forall \lambda \in [\lambda_1, \infty]\)
we have \(\bar \beta_{s^*}(\lambda) = \bar \beta_{{s}(\lambda)}(\lambda)\). Under {\bf A0}, we know that
\(\forall 0 < \lambda \leq \lambda_1\), \(i_1^*\) will remain selected and that the relative order of
\(i_1^*\) in the set of variables \(G_{\Gamma(i_1^*)}\), as induced by the magnitude of their coefficients in
\(\bar \beta_{{s}(\lambda_1)}(\lambda_1)\) will not change. Using this and the general form of \(s^* \in
\mathcal{S}_{\theta,\mathcal{G}}\) given by Proposition~\ref{pro:decomposition}, we can modify \(s(\lambda)\)
in a way that is consistent with Proposition~\ref{pro:underest}. Specifically, if we let \(t = 1\), then the
current active set is \(A_t = \{i: \left|X_i^\top (y - X \bar
\beta_{s(\lambda_t)}(\lambda_t))\right|/s_i(\lambda_t) = \lambda_t\}\). We may apply the following generic update
to \({s}(\lambda)\) so that at \(\lambda_t^-\) (i.e., for a value of \(\lambda\) infinitesimally smaller than
\(\lambda_t\)) it satisfies
\begin{align}
{s}_j(\lambda_t^-) &=\!\left\{\!\!\!\!\begin{array}{cl}
{s}_j(\lambda_t)\!+\!\frac{\theta_{\Gamma(i_t^*)}}{|G_{\Gamma(i_t^*)}| - 1} &\!\!\!j \in \{A_t^c \cap
G_{\Gamma(i_t^*)}\}\\
{s}_j(\lambda_t) &\!\!\!\mbox{o.w.}
\end{array}\right.\!\!\!.\label{eq:adaptation}
\end{align}
Notice that the change leaves the path \(\bar \beta_{{s}(\lambda_t)}(\lambda_t)\) continuous in the
neighborhood of \(\lambda_t\). RepLasso then continues to decrease \(\lambda = \lambda_1 \downarrow 0\), again
keeping \({s}(\lambda) = {s}(\lambda_1^-)\) constant and producing solutions \(\bar
\beta_{{s}(\lambda)}(\lambda)\) along the way, until a point \(\lambda_{2} > 0\) is reached when a new variable is
selected by \(\bar \beta_{{s}(\lambda)}(\lambda)\). Because \({s}(\lambda)\) was kept constant for \(\lambda
\in [\lambda_2, \lambda_1^-]\), this can be achieved by a straightforward modification of the Lasso homotopy
method\footnote{Specifically, where the Lasso homotopy method traces out {\it
    equiangular} directions, the RepLasso follows {\it skew-angular} directions (given in stage 2), with the
  angle skew determined by the weights \(s_A(\lambda)\).}. As before, we know from our update of
\({s}(\lambda)\) and Proposition~\ref{pro:underest} that with probability 1, \(\forall \lambda \in [\lambda_2,
  \lambda_1]\) we have \(\bar \beta_{s^*}(\lambda) = \bar \beta_{{s}(\lambda)}(\lambda)\). At this point, {\bf
  A0} and Proposition~\ref{pro:decomposition} again allow us to update \({s}(\lambda)\) using
Eq.~(\ref{eq:adaptation}) with \(t = 2\). RepLasso continues sweeping out \(\lambda\) in this fashion until
some final value \(\lambda_T > 0\) is reached. By the time the algorithm has completed, we know that with probability 1, \(\forall
\lambda \in [\lambda_T, \infty]\) we have \(\bar \beta_{s^*}(\lambda) = \bar
\beta_{{s}(\lambda)}(\lambda)\). The final claim follows immediately. 
\end{proof}

When {\bf A0} does not hold, we can apply Proposition~\ref{pro:underest} to~(\ref{eq:nonconvex_equiv}) to see
that RepLasso will generally still recover global minima of~(\ref{eq:nonconvex}) for large \(\lambda >
0\). Indeed, if RepLasso adds variables one by one, the first variable selected by RepLasso is also the first
selected by~(\ref{eq:nonconvex}). Regardless of whether {\bf A0} holds, Section~\ref{sec:analysis} shows
strong results for RepLasso relative to the \(\ell_1\) relaxation of~(\ref{eq:nonconvex}) (i.e., the Lasso).

%% file: analysis.tex
\section{Comparing RepLasso and Lasso}\label{sec:analysis}
In this section we show several results irrespective of whether {\bf A0} holds, but assuming
that \(\mathcal{G}\), \(\beta^*\) satisfy certain conditions. Before continuing, we briefly outline some
notation. Let \(X_j\) be the column \(j\) of \(X\) and \(X_A\) a matrix which consists of the columns indexed
by \(A\). Let the support set of \(\beta^*\) be \(S \triangleq S(\beta^*)\). Denote the signed support of
\(\beta^*\) by \(S_{\pm} = S_\pm(\beta^*)\), where
\begin{align}
  S_\pm(\beta_i) \triangleq \left\{\begin{array}{rl}
  +1 & \mbox{if}~\beta_i > 0\\
  -1 & \mbox{if}~\beta_i < 0\\
  0 & \mbox{o.w.}\end{array}\right..
\end{align}
Our analysis in this section relies on various subsets of the following assumptions
\vspace{2mm}
\begin{itemize}
\itemsep0em
\item[{\bf A1}:] \(\forall G \in \mathcal{G}, |\{i \in G: \beta_i^* \neq 0\}| \leq 1\)
\item[{\bf A2}:] \(\forall A\subset S\) and \(u_A\) the equiangular vector in Eq.~(2.6) of~\cite{efron04lars},  \(\nexists j
  \in A^c, |X_A^\top u_A| = |X_j^\top u_A| {\bf 1}\)\todo{Maybe we don't even need this. Would this imply that
    \(X_A^\top X_A\) is singular?}
\item[{\bf A3}:] \(X_S^\top X_S\) is invertible
\item[{\bf A4}:] Following~\cite{wainwright09sharp,zhao2006model}, define
\begin{align*}
\mu_j &\triangleq X_j^\top X_S (X_S^\top X_S)^{-1} \mbox{sgn}(\beta_S^*) \\
\gamma_i &\triangleq e_i^\top \left(X_S^\top X_S\right)^{-1}n.
\end{align*}
We have 
\(\forall j \in
  S^c, |\mu_j| < 1\), \(\forall i \in
  S, \mbox{sgn}(\beta_i^*) \gamma_i > 0\).
\end{itemize}
Assumption {\bf A1} formalizes that \(\beta^*\) is nonzero on at most a few elements (in this case one) of
each group of \(\mathcal{G}\). Assumption {\bf A2} ensures that the active set estimated by the algorithm
in~\cite{efron04lars} matches the support set and holds with probability 1 if \(X\) has spherical and
absolutely continuous distribution. Assumption {\bf A3} is a zeroth-order condition necessary for
identifiability. Assumption {\bf A4} ensures that the Lasso has any chance of recovering the signed support of
\(\beta^*\).

Many analyses of the Lasso focus on its support recovery properties. The following theorem compares 
RepLasso against Lasso in terms of this measure.
\begin{thm}[Support Subset Recovery]\label{thm:support_lasso}
Assume that {\bf A1--2} hold. Denote by \(\hat\beta(\lambda)\) and \(\bar\beta(\lambda)\) the
Lasso and RepLasso solutions for penalty parameter \(\lambda\). Given \(X, y\), we have for any \(\lambda_{\min} >
0\)
\begin{align*}
  \forall \lambda \geq \lambda_{\min}~S(\hat\beta(\lambda)) \subseteq S 
  \implies 
  \forall
  \lambda \geq \lambda_{\min}~S(\bar\beta(\lambda)) \subseteq S.
\end{align*}
\end{thm}
{\it Proof sketch.} Suppose \(\forall \lambda \geq \lambda_{\min}~S(\hat\beta(\lambda)) \subseteq S\).  For
\(t' \leq t\), let \(\hat A_{t'}\) and \(\bar A_{t'}\) be the active sets of the Lasso and RepLasso at iteration
\(t'\) until \(\lambda_{\min}\) is reached. By {\bf A2} we can show that \(\forall t' \leq t, \hat A_{t'}
\subseteq S\). Since \(s(|\!|X^\top y|\!|_\infty) = {\bf 1}\) we know that \(\hat A_{1} = \bar A_{1} \subseteq
S\). It then follows from \(\hat A_{2} \subseteq S\), assumption {\bf A1} and the construction of
\(s(\lambda)\) that we also have \(\hat A_{2} = \bar A_{2} \subseteq S\). Iterating this argument over \(t'
\leq t\), we can then show that \(\forall t' \leq t, \bar A_{t'} \subseteq S\) from which it follows that
\(\forall \lambda \geq \lambda_{\min}~S(\bar\beta(\lambda)) \subseteq S\).

In many situations we are not only interested in recovering a subset of the true support, but the signed
support of \(\beta^*\). The previous result can be strengthened to cover this case.
\begin{thm}[Signed Support Recovery]\label{thm:signed_support_lasso}
Assume that {\bf A1--4} hold. Denote by \(\hat\beta(\lambda)\) and \(\bar\beta(\lambda)\) the Lasso and RepLasso
solutions using penalty parameter \(\lambda\). For any \(\lambda_{\min} > 0\), we have with probability 1 over
an absolutely continuous distribution on noise \(w\)
\begin{align*}
\forall \lambda \geq \lambda_{\min}~S(\hat\beta(\lambda))
\subseteq S, S_\pm(\hat \beta(\lambda_{\min})) = S_\pm 
\implies
\forall \lambda \geq \lambda_{\min}~S(\bar\beta(\lambda)) \subseteq S, S_\pm(\bar \beta(\lambda_{\min})) = S_\pm.
\end{align*}
\end{thm}
{\it Proof sketch.} Suppose \(\forall \lambda \geq \lambda_{\min}~S(\hat\beta(\lambda)) \subseteq S,
S_\pm(\hat \beta(\lambda_{\min})) = S_\pm\). For \(t' \leq t\), let \(\hat A_{t'}, \bar A_{t'}\) be the active
sets of Lasso/RepLasso until \(\lambda_{\min}\) is reached. As in Theorem~\ref{thm:support_lasso}, we conclude
from {\bf A2} that \(\forall t' \leq t\), \(\hat A_{t'} \subseteq S\), which then tells us via {\bf A1} that
\(\forall t' \leq t, \bar A_{t'} \subseteq S\) and so \(\forall \lambda \geq \lambda_{\min},
S(\bar\beta(\lambda)) \subseteq S\). Furthermore, by {\bf A1} and the construction of \(s(\lambda)\), we know
\(\forall \lambda \geq \lambda_{\min}\) that \(s_S(\lambda) = {\bf 1}\) and \(s_{S^c}(\lambda) \geq {\bf 1}\). Utilizing
          {\bf A3--4} and Lemma~1 of Wauthier et al.~\cite{wauthier2013plasso} (which holds with probability 1
          over noise \(w\)), we can then argue that with probability 1, \(S_\pm(\bar \beta(\lambda_{\min})) =
          S_\pm\).

\paragraph{Consequences for other Methods.}
Besides the Lasso, Theorems~\ref{thm:support_lasso} and~\ref{thm:signed_support_lasso} also apply to many
related algorithms that pre-process the data \(X, y\) in some way, prior to running the Lasso on the modified
data. Instances of these algorithms are, for example, the Adaptive Lasso~\cite{zou2006adaptive} and various
Preconditioned Lasso algorithms~\cite{huang11corrsift,jia12whitened,paul08plasso}. Indeed, if the relevant
assumptions {\bf A1--4} hold, the result is even true for \(\ell_1\) regularized minimization of quadratic
approximations to logistic regression as proposed in~\cite{lee2006efficient}. We will empirically highlight
this property in Section~\ref{sec:results}.


\begin{figure*}[t]
\begin{center}
\begin{tabular}{c}
\subfigure[\(\rho \approx 0.1\)]{
  \includegraphics[width=6.3cm]{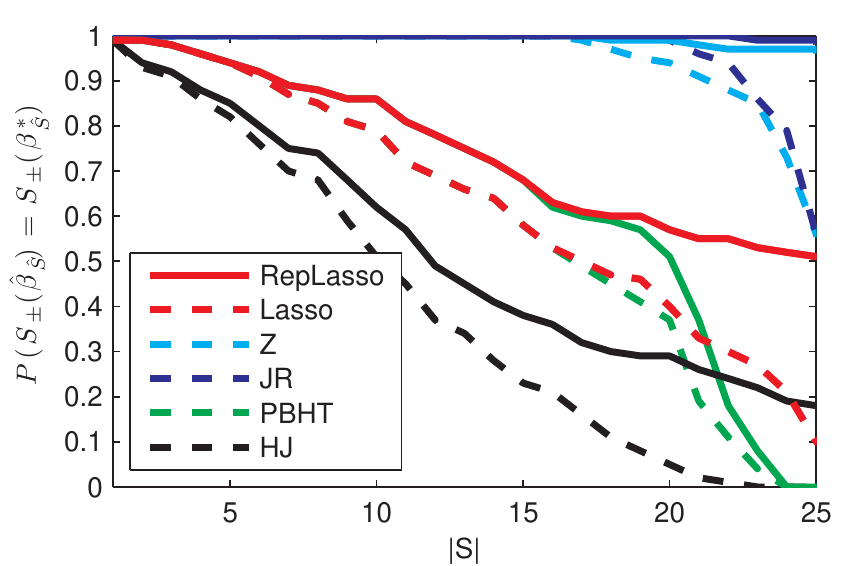}\label{subfig:pairsrho01}
}\\
\subfigure[\(\rho \approx 0.5\)]{
  \includegraphics[width=6.3cm]{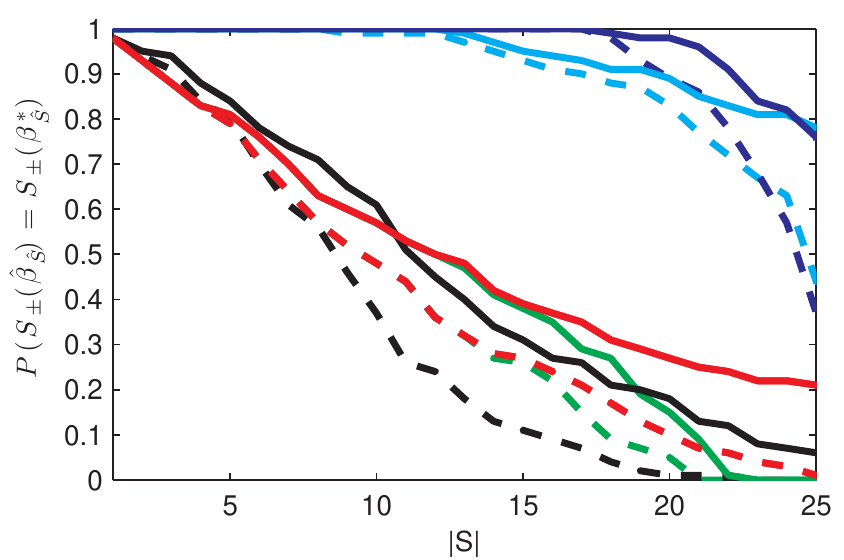}\label{subfig:pairsrho05}
}
\end{tabular}
\begin{tabular}{c}
  \subfigure[Marginal MLRT]{
    \includegraphics[width=6.7cm]{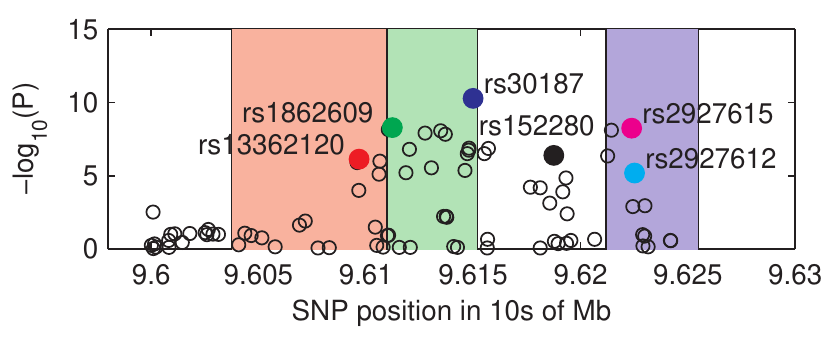}\label{subfig:asgwas}
  }\\
  \subfigure[IRLS-Lasso]{
    \includegraphics[width=6.7cm]{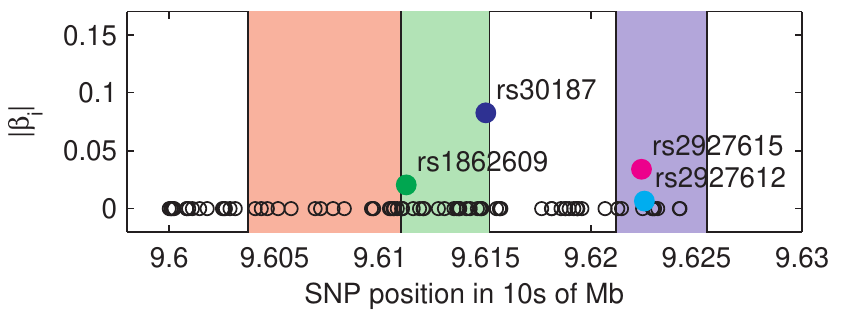}\label{subfig:aslasso}
  }\\
  \subfigure[IRLS-RepLasso]{
    \includegraphics[width=6.7cm]{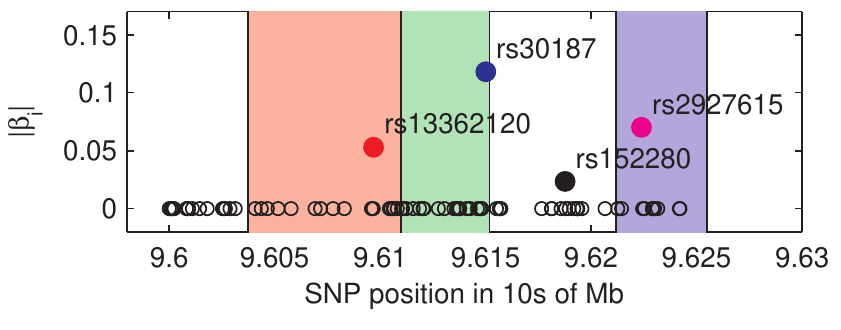}\label{subfig:asreplasso}
  }
\end{tabular}

\end{center}
\caption{~\subref{subfig:pairsrho01},~\subref{subfig:pairsrho05} Results on synthetic data with \(n = 150, p =
  50, \mathcal{G} = \mbox{groups of 2}\)
  and within-group correlations \(\rho\). We show the empirical probability that a subset of the correct {\it
    signed} support is recovered as a function of support size. For each base-method we show two curves,
  grouped by color. The performance of the existing Lasso variant (i.e., Lasso, Z, JR, PBHT, HJ) is shown as
  dashed curve; the performance of the algorithm with the Lasso replaced by RepLasso is shown in
  solid (see text for details).~\subref{subfig:asgwas} A GWAS Manhattan plot for ankylosing spondylitis on a region of chromosome
  5. The \(x\)-axis gives SNP position; circles indicate the \(-\log_{10}(p)\)-values of marginal association
  tests. The red, green and blue shaded regions indicate the CAST, ERAP1 and ERAP2 genes respectively. Solid
  circles highlight the SNPs that were chosen by the methods below.~\subref{subfig:aslasso} The IRLS method of
  Lee et al.~\cite{lee2006efficient} with Lasso. Circles indicate the magnitudes of estimated \(\beta\)
  coefficients. Solid circles indicate the first 4 SNPs that are chosen by Lasso. The method selects multiple
  SNPs from the same gene.~\subref{subfig:asreplasso} The RepLasso avoids this if \(\theta\) is chosen large
  enough.}
\label{fig:results}
\end{figure*}

\paragraph{A Lars-like Variation.} We note at this point that the Lars algorithm~\cite{efron04lars} is a special case of
RepLasso if we set \(\theta = {\bf 0}\) and force \(L = 0\). If we only force \(L = 0\) but allow
\(\theta \neq {\bf 0}\), then the resulting algorithm can be seen as a generalization of Lars. We analyze its
performance in the Supplementary Material and show a similar result for its support recovery behavior.

%% file: results.tex
\section{Results}\label{sec:results}
\paragraph{Synthetic Data.} To underline the findings of Section~\ref{sec:analysis} we  focus on a set of experiments
which analyses the probability of correctly recovering a subset of the correct {\it signed} support. We fix
\(\mathcal{G}, \beta^*\) so that {\bf A1} holds. Conditioned on \(\mathcal{G}\) we also sample \(X\) with
unit-length columns that are independent between groups \(G \in \mathcal{G}\) but exhibit some correlation
\(\rho\) within groups. Given \(X, \beta^*\), we generate \(y\) according to Eq.~(\ref{eq:genmodel}), with
\(\sigma^2 = 0.2^2\). In Figures~\ref{subfig:pairsrho01} and~\ref{subfig:pairsrho05} we investigate the
performance of the RepLasso (solid red) and the Lasso (dashed red). Notice that the curve for RepLasso lies
above that of Lasso, giving empirical support to Theorems~\ref{thm:support_lasso}
and~\ref{thm:signed_support_lasso}. Additionally, we evaluate the performance of four other methods that
solve a standard Lasso problem after pre-processing the data \(X, y\) in some way. For each algorithm we show
two curves, grouped by colors: the original method is shown as dashed curve, the method with the Lasso
replaced by RepLasso as solid curve. The four methods are: (1) the Adaptive Lasso of
Zou~\cite{zou2006adaptive} ({\it Z})\fixme{rescales X}; (2) the ``Whitened'' Lasso of Jia and
Rohe~\cite{jia12whitened} ({\it JR}); (3) the Preconditioned Lasso of Paul et al.~\cite{paul08plasso} ({\it
  PBHT}); and (4) Correlation Sifting of Huang and Jojic~\cite{huang11corrsift} ({\it HJ}). The results in
Figures~\ref{subfig:pairsrho01} and~\ref{subfig:pairsrho05} highlight that using RepLasso as a drop-in
replacement these algorithms can also be improved. Results are similar when \(\mathcal{G}\) contains
larger groups or when \(n \ll p\).\footnote{If \(n < p\) we let the Adaptive Lasso scale columns of \(X\) by
  univariate regression coefficients.}

\paragraph{GWAS application.} A second experiment considers the application of the Lasso to a Genome-Wide
Association Study (GWAS). A GWAS hopes to find Single Nucleotide Polymorphisms (SNPs) that are associated with
disease status. Our focus is on the disease ankylosing spondylitis and a region of chromosome 5, where
susceptibility SNPs had been previously reported~\cite{burton2007association}. The mainstream GWAS methodology
tests each form a large set of SNPs {\it marginally} for association using a maximum likelihood ratio test
(MLRT) and plots the resulting \(p\)-values on a ``Manhattan plot'', as in Figure~\ref{subfig:asgwas}. Due to
linkage disequilibrium, many small \(p\)-values lie close to each other. Alternatively, a penalized logistic
regression could also be used to regress the SNPs onto disease status, which would then highlight interesting
SNPs by the magnitudes of the learned regression coefficients. Lee et al.~\cite{lee2006efficient} have
proposed an IRLS strategy for estimating an \(\ell_1\) constrained logistic regression by solving a Lasso
problem on a quadratic approximation of the logistic objective. The magnitudes of the first four regression
coefficients estimated by this method are shown in Figure~\ref{subfig:aslasso}. As can be seen, two pairs of
selected SNPs lie near each other in two genes. A researcher might wish to discourage the Lasso from choosing
multiple SNPs from the same gene. The RepLasso is ideally suited to this task. Given the gene partition
\(\mathcal{G}\) (in this case by CAST, ERAP1 and ERAP2 genes) we can replace the Lasso in the IRLS algorithm
by the RepLasso and produce a different parameter estimate. If \(\theta\) is large (e.g., 20 for each 
group), RepLasso avoids selecting multiple SNPs from the same gene, as seen in
Figure~\ref{subfig:asreplasso}. These SNPs may be worthy of further study.


%% file: conclusion.tex
\section{Conclusion}\label{sec:conclusion}
\vspace{-2mm}
In this paper we presented a homotopy-style algorithm that approximates an underlying nonconvex
problem by producing a suitable sequence of surrogates that locally approximate
\(\Omega_{\theta,\mathcal{G}}(\cdot)\) well. The Lasso approach, in comparison, revolves around finding a
single global surrogate that often approximates \(\Omega_{\theta,\mathcal{G}}(\cdot)\) poorly. As shown by
Theorem~\ref{thm:replars_optimality}, our method will in certain cases sweep out a global minima path
of~(\ref{eq:nonconvex}). Further, we showed in Section~\ref{sec:analysis} that even though RepLasso may not
exactly solve~(\ref{eq:nonconvex}) in general, in relevant regression problems RepLasso will not do worse than
the Lasso and in practice often outperforms it.

Several extensions
can be considered. Firstly, we defined \(\Omega_{\theta, \mathcal{G}}(\cdot)\) as a sum over pairs of
variables in groups of the partition \(\mathcal{G}\). More flexible constraint functions could potentially be
defined if the sum is allowed to be over an arbitrary set of pairs. Secondly, our overall strategy was to
decompose the nonconvex constraint balls induced by \(\Omega_{\theta, \mathcal{G}}(\cdot)\) as a union of
simpler, convex balls. In this work the constraint function \(\Omega_{\theta, \mathcal{G}}(\cdot)\) gave rise
to a union of weighted \(\ell_1\) balls. This motivates a more direct definition of nonconvex constraint balls
as a union of convex balls. For instance, one could consider unions of weighted \(\ell_\infty\) balls or a mix
of weighted \(\ell_\infty\) and weighted \(\ell_1\) balls. So long as these convex building blocks are
consistent with~\cite{rosset2007piecewise} it should still be possible to efficiently compute local minima
paths segments as demonstrated in this paper. Thirdly, it would be interesting to see whether results such as
in Loh and Wainwright~\cite{loh13mest} could be extended to argue for statistical consistency of the RepLasso
in cases where local minima paths are produced. An important ingredient of such an analysis will be that
\(\Omega_{\theta,\mathcal{G}}(\beta)\) is not ``too'' nonconvex, which might hold if \(\theta\) is
sufficiently small.

%% file: acknowledgements.tex
\subsection*{Acknowledgments}
We thank Francis Bach and Alexander Young for helpful comments and Nebojsa Jojic for early input out of which
this research grew.

%% file: supplementary.tex
\section{Proofs of Section 3}
Recall that given a partition \(\mathcal{G} = \{G_1, \ldots, G_g\}\) of \(\{1, \ldots, p\}\) without singleton or empty
sets, and a vector \(\theta = (\theta_1, \ldots, \theta_g) \geq 0\), we defined
\begin{align}
\Omega_{\theta,\mathcal{G}}(\beta)\!&=\!\!\!\!\!\sum_{i<j \in G_{g'} \in \mathcal{G}}\!\!\!\frac{\omega_{\theta_{g'}}(\beta_i, \beta_j)}{|G_{g'}| -
  1}~~~~~~\omega_{\theta_{g'}}(\beta_i, \beta_j) = \min(|\beta_i|,|\beta_j|)(1 + \theta_{g'}) +
\max(|\beta_i|,|\beta_j|). 
\end{align}
Let \(B_{\theta, \mathcal{G}}(\tau) =
\left\{\beta \in \mathbb{R}^p: \Omega_{\theta, \mathcal{G}}(\beta) \leq \tau\right\}\) be the induced
constraint balls and \(\Gamma(i) \in \{1, \ldots, g\}\) the (unique) group so that \(i \in G_{\Gamma(i)}\).
\subsection{Proof of Proposition 1}
\begin{labelpro}{1}[Union Decomposition]\label{pro:decomposition_app}
Let the partition be \(\mathcal{G} = \{G_1, \ldots, G_g\}\) and the parameter \(\theta = (\theta_1, \ldots,
\theta_g) \geq 0\). There is a finite set \(\mathcal{S}_{\theta,\mathcal{G}} \subset \mathbb{R}^p\) of vectors
\(s \geq {\bf 1}\), so that for any \(\tau > 0\)
\begin{align}
B_{\theta, \mathcal{G}}(\tau) = \bigcup_{s \in \mathcal{S}_{\theta,\mathcal{G}}} B_{s}(\tau).
\end{align}
Define \(\Pi_{g'}\) to be all permutations \(\pi_{g'}\) of the elements in \(G_{g'}\) and let \(\Pi_{\mathcal{G}} =
\times_{g' = 1}^g \Pi_{g'}\) be their cross-product, whose elements \(\pi \in \Pi_{\mathcal{G}}\) are \(g\)-tuples of
permutations \(\pi = (\pi_1, \ldots, \pi_g)\). For some \(\pi \in \Pi_{\mathcal{G}}\), denote by \(\pi_{\Gamma(i)}(i) \in
\{1 , \ldots, |G_{\Gamma(i)}|\}\) the position of \(i \in G_{\Gamma(i)}\) in permutation \(\pi_{\Gamma(i)}\).  We have
\begin{align}
\mathcal{S}_{\theta,\mathcal{G}} &= \cup_{\pi \in \Pi_{\mathcal{G}}} \{s_\pi\}\\
s_{\pi, i} &= 1 + (\pi_{\Gamma(i)}(i)-1) \frac{\theta_{\Gamma(i)}}{|G_{\Gamma(i)}| - 1}~\forall i = 1, \ldots, p.
\end{align}
\end{labelpro}
\begin{proof}
We first show \(B_{\theta, \mathcal{G}}(\tau) \subseteq \bigcup_{s \in \mathcal{S}_{\theta,\mathcal{G}}} B_{s}(\tau)\). Consider some \(\beta
\in B_{\theta, \mathcal{G}}(\tau)\) and let \(\pi = (\pi_1, \ldots, \pi_g)\) be a tuple of permutations (not
necessarily unique) induced by sorting the elements \(|\beta_i|\) within each group specified by
\(\mathcal{G}\) so that for each group index \(g'\) we have
\begin{align}
|\beta_{\pi_{g'}^{-1}(1)}| \geq \ldots \geq |\beta_{\pi_{g'}^{-1}(|G_{g'}|)}|.
\end{align}
By construction of
\(\Omega_{\theta,\mathcal{G}}(\cdot)\), \(\beta\) lies in the set
\begin{align}
&~~~~\left\{\beta' \in \mathbb{R}^p: \sum_{g' = 1}^g \sum_{i=1}^{|G_{g'}|} \left(\frac{(|G_{g'}| - i) + (i -
    1)(1 + \theta_{g'})}{|G_{g'}| - 1}\right) |\beta'_{\pi_{g'}^{-1}(i)}| \leq
\tau\right\} \\
&= \left\{\beta' \in \mathbb{R}^p: \sum_{g' = 1}^g \sum_{i=1}^{|G_{g'}|} \left(1 + \frac{(i - 1)\theta_{g'}}{|G_{g'}| - 1}\right) |\beta'_{\pi_{g'}^{-1}(i)}| \leq
\tau\right\} \\
&= \left\{\beta' \in \mathbb{R}^p: \sum_{g' = 1}^g\sum_{i \in G_{g'}} \left(1 + \left(\pi_{g'}(i) -
1\right)\frac{\theta_{g'}}{|G_{g'}| - 1}\right) |\beta'_{i}| \leq
\tau\right\} \\
&= \left\{\beta' \in \mathbb{R}^p: \sum_{i = 1}^p \left(1 + \left(\pi_{\Gamma(i)}(i) - 1\right)\frac{\theta_{\Gamma(i)}}{|G_{\Gamma(i)}| - 1}\right) |\beta'_{i}| \leq
\tau\right\}  = B_{s_\pi}(\tau)
\end{align}
We therefore conclude that \(B_{\theta, \mathcal{G}}(\tau) \subseteq \bigcup_{s \in
  \mathcal{S}_{\theta,\mathcal{G}}} B_{s}(\tau)\), with
\begin{align}
\mathcal{S}_{\theta,\mathcal{G}} &= \cup_{\pi \in \Pi_{\mathcal{G}}} \{s_\pi\}\\
s_{\pi, i} &= 1 + (\pi_{\Gamma(i)}(i)-1) \frac{\theta_{\Gamma(i)}}{|G_{\Gamma(i)}| - 1}~\forall i = 1, \ldots, p.
\end{align}
For the other direction, suppose that \(\beta \in B_{s_{\tilde \pi}}(\tau)\) for some arbitrary tuple of
permutations \(\tilde \pi \in \Pi_{\mathcal{G}}\), which means that
\begin{align}
\sum_{i = 1}^p \left(1 + \left(\tilde\pi_{\Gamma(i)}(i) - 1\right)\frac{\theta_{\Gamma(i)}}{|G_{\Gamma(i)}| - 1}\right) |\beta_{i}| \leq \tau.
\end{align}
Then notice that if \(\pi\) is a (not necessarily unique) tuple of permutations induced by ordering elements \(|\beta_i|\) within
groups, we have, by arguing from pairwise swaps within groups that take \(\tilde \pi\) to \(\pi\), that
\begin{align}
\sum_{i = 1}^p \left(1 + \left(\pi_{\Gamma(i)}(i) - 1\right)\frac{\theta_{\Gamma(i)}}{|G_{\Gamma(i)}| - 1}\right) |\beta_{i}| \leq
  \sum_{i = 1}^p \left(1 + \left(\tilde\pi_{\Gamma(i)}(i) - 1\right)\frac{\theta_{\Gamma(i)}}{|G_{\Gamma(i)}| - 1}\right) |\beta_{i}| \leq
  \tau,
\end{align}
and so \(\beta \in B_{\theta,
  \mathcal{G}}(\tau)\). It follows that \( \bigcup_{s \in \mathcal{S}_{\theta,\mathcal{G}}} B_{s}(\tau) \subseteq B_{\theta,
  \mathcal{G}}(\tau)\) and so \(B_{\theta, \mathcal{G}}(\tau) = \bigcup_{s \in \mathcal{S}_{\theta,\mathcal{G}}} B_{s}(\tau)\).  


\end{proof}
\newpage

\subsection{Proof of Proposition 2}
Recall that we are considering the nonconvex optimization problem
\begin{align}
\beta(\tau) &\in 
\underset{\beta\in \mathbb{R}^p}{\mbox{argmin}} J_\tau(\beta) \tag{$P1$}\label{eq:nonconvex_app}\\
&= \underset{\beta\in \mathbb{R}^p}{\mbox{argmin}} \left\{\begin{array}{cl}
\frac{1}{2n}\left|\!\left|y - X\beta\right|\!\right|_2^2 &\mbox{if}~~\beta \in B_{\theta, \mathcal{G}}(\tau)\\
\infty & \mbox{o.w.} 
\end{array}
\right.\!\!\!\!.\nonumber
\end{align}

\begin{labelpro}{2}[Local Piecewise Linearity]
Suppose \(X\) has absolutely continuous distribution and that \(\exists \tau' > 0\)
s.t. \(\nexists \beta \in B_{\theta, \mathcal{G}}(\tau')\) which is a minimum of \(|\!|y -
X\beta|\!|_2^2\). Let \(\tau_{\max}\) be the supremum over these \(\tau'\). The set of local minima of \(J_\tau(\cdot)\) in~(\ref{eq:nonconvex_app}) with \(\tau \in (0,
\tau_{\max})\) is w.p. 1 a finite union of piecewise linear paths, each path indexed by \(\tau\)
  and lying on a ball \(B_{s}(\tau), s \in \mathcal{S}_{\theta,\mathcal{G}}\). 
\end{labelpro}
\begin{proof}
Given the assumptions, for all \(\tau \in (0, \tau_{\max})\), the elements \(\beta\) on the boundary of
\(B_{\theta, \mathcal{G}}(\tau)\) satisfy \((y - X\beta)^\top X \neq 0.\) For each \(\tau \in (0,
\tau_{\max})\), let \(\mathcal{M}_{\theta, \mathcal{G}}(\tau)\) be the set of local minima of
\(J_{\tau}(\cdot)\). Let the set \(\mathcal{S}_{\theta,\mathcal{G}}\) be defined as in
Proposition~\ref{pro:decomposition_app}: For \(\Pi_{\mathcal{G}}\) the set of \(g\)-tuples of permutations induced
by \(\mathcal{G}\),
\begin{align}
\mathcal{S}_{\theta,\mathcal{G}} &= \cup_{\pi \in \Pi_{\mathcal{G}}} \{s_\pi\}\\
s_{\pi, i} &= 1 + (\pi_{\Gamma(i)}(i)-1) \frac{\theta_{\Gamma(i)}}{|G_{\Gamma(i)}| - 1}~\forall i = 1, \ldots,
p.\label{eq:lemsconstruction}
\end{align}
For some \(s_\pi \in \mathcal{S}_{\theta,\mathcal{G}}\), define \(\mathcal{M}_{s_\pi}(\tau)\) to be the
solution to~(\ref{eq:nonconvex_app}) with \(B_{\theta, \mathcal{G}}(\tau)\) replaced by
\(B_{s_\pi}(\tau)\). For each \(s_\pi \in\mathcal{S}_{\theta,\mathcal{G}}\) the ball \(B_{s_\pi}(\tau)\)
corresponds to a weighted \(\ell_1\) norm, and if \(X\) is drawn from an absolutely continuous distribution,
then the solution \(\mathcal{M}_{s_\pi}(\tau)\) is with probability 1 unique on \((0,
\tau_{\max})\)~\cite{tibshirani2013lasso}. Additionally, the result of Rosset and
Zhu~\cite{rosset2007piecewise} shows that the resulting regularization path \(\mathcal{M}_{s_\pi}(\tau)\) is
piecewise linear on \((0, \tau_{\max})\). Due to the union decomposition of
Proposition~\ref{pro:decomposition_app}, it follows immediately that \(\mathcal{M}_{\theta, \mathcal{G}}(\tau)
\subseteq \bigcup_{s_\pi \in \mathcal{S}_{\theta,\mathcal{G}}} \mathcal{M}_{s_\pi}(\tau)\) for \(\tau \in (0,
\tau_{\max})\). However, we seek not a superset of \(\mathcal{M}_{\theta, \mathcal{G}}(\tau)\), but a
characterisation as a union of paths on the boundaries of weighted \(\ell_1\) balls. That, is we seek a set
\(P \subseteq \Pi_{\mathcal{G}}\) so that \(\mathcal{M}_{\theta, \mathcal{G}}(\tau) = \bigcup_{\pi \in P}
\mathcal{M}_{s_\pi}(\tau)\) for \(\tau \in (0, \tau_{\max})\). The existence of such a set \(P\) can be
guaranteed if for any \(s_\pi \in \mathcal{S}_{\theta,\mathcal{G}}\), \(\mathcal{M}_{s_\pi}(\tau)\) either
lies \(\forall \tau \in (0, \tau_{\max})\) in the interior of \(B_{\theta, \mathcal{G}}(\tau)\) or it lies
\(\forall \tau \in (0, \tau_{\max})\) on the boundary of \(B_{\theta, \mathcal{G}}(\tau)\). To show this, we
show that for \(\tau \in (0, \tau_{\max})\) no local minimum in \(\mathcal{M}_{\theta, \mathcal{G}}(\tau)\)
lies at a concave kink of \(B_{\theta, \mathcal{G}}(\tau)\) (which are the points where a path would switch
from being in the interior to being on the boundary or vice versa).


Suppose then (for the purpose of deriving a contradiction) that for
  some \(\tau \in (0, \tau_{\max})\), we have that \(\beta\) is a local
    minimum in \(\mathcal{M}_{\theta,
  \mathcal{G}}(\tau)\) that lies at one of the concave kinks of \(B_{\theta, \mathcal{G}}(\tau)\). If \(\beta\) lies at a
    concave kink, then since \(\tau_{\max} > 0\), we know that for at least two elements \(i\neq j \in G
    \in \mathcal{G}\), \(\beta_i \neq 0, \beta_j \neq 0\). For if only a single element \(\neq 0\), then
    we lie at one of the points of \(B_{\theta,\mathcal{G}}(\tau)\) and if the only two nonzero elements lie
    in different groups, \(\beta\) cannot lie at a concave kink. Specifically, the concave kink is identified
    by sets of indices \(i\) in a group \(G \in \mathcal{G}\) so that the corresponding \(\beta_i \neq 0\)
    have identical magnitude. The vector \(\beta\) induces a set \(\Sigma \subseteq \Pi_{\mathcal{G}}\) of
    \(g\)-tuples of permutations \(\sigma\) by sorting \(|\beta_i|\) by their magnitudes within each group
    \(G\in\mathcal{G} = \{G_1, \ldots, G_g\}\) (with tie-breaking). We know that for each \({\sigma} \in
    \Sigma\), \(|\!|\mbox{diag}(s_{\sigma})\beta|\!|_1 = \tau\), that is, \(\beta\) lies on the boundary of
    \(B_{s_{\sigma}}(\tau)\). Each \({\sigma}\) thus corresponds to an active constraint on \(\beta\). Since
    we can think of \(\beta\) as a local minimum of \(|\!|y - X\beta|\!|_2^2\), subject to either of these
    (convex) constraints, we have by convexity for any \({\sigma} \in \Sigma\) a subgradient
    vector \(z_\sigma \in \partial |\!|\beta|\!|_1\) and a constant \(\lambda_\sigma\) so that
\begin{align}
  (y - X\beta)^\top X  &= \lambda_\sigma \mbox{diag}(z_\sigma)s_{{\sigma}}.\label{eq:optcond}
\end{align}
Because there are at least two elements \(i\neq j \in G \subseteq \mathcal{G}\) with \(|\beta_i| = |\beta_j|
\neq 0\) we know that \(\forall \sigma \in \Sigma\), \(z_{\sigma,i} = \mbox{sgn}(\beta_i), z_{\sigma,j} = \mbox{sgn}(\beta_j)\), which
implies that \(z_{\sigma,i}s_{{\sigma},i} \neq 0, z_{\sigma,j}s_{{\sigma},j} \neq 0\). Additionally, by
construction \((y - X\beta)^\top X \neq 0\) and so we know \(\lambda_\sigma \neq 0\).  By the construction of
\(s_{\pi}\) in Eq.~(\ref{eq:lemsconstruction}), we know that \(\exists \sigma_1 \neq \sigma_2 \in \Sigma\), so
that \(s_{{\sigma}_1}\) and \(s_{{\sigma}_2}\) differ only on elements \(i,j\). However Eq.~(\ref{eq:optcond})
then cannot simultaneously hold unless \(\lambda_\sigma = 0\) and \((y - X\beta)^\top X = 0\) which we ruled
out earlier. Thus we have a contradiction and so the assumption that \(\beta\) lies at a concave kink must be
wrong.



Because local minima in \(\mathcal{M}_{\theta, \mathcal{G}}(\tau)\) never lie at concave kinks of \(B_{\theta, \mathcal{G}}(\tau)\) for \(\tau \in (0,
\tau_{\max})\), we know that for
  each local minimum path on \((0, \tau_{\max})\), there is a \(\pi \in \Pi_{\mathcal{G}}\) so that the path
    lies on \(B_{s_\pi}(\tau)\). That is, there is some nonempty subset \(P \subseteq \Pi_{\mathcal{G}}\) so
    that \(\mathcal{M}_{\theta, \mathcal{G}}(\tau) = \bigcup_{\pi \in P} \mathcal{M}_{s_\pi}(\tau)\) is a
    union of piecewise linear paths.
\end{proof}
\newpage

\section{Proofs of Section 4}

\subsection{Proof of Proposition 3}
Recall that we are considering the surrogate problem 
\begin{align}
\bar \beta(\lambda) &\in \mbox{argmin}_{\beta\in \mathbb{R}^p} \frac{1}{2n}\left|\!\left|y -
X\beta\right|\!\right|_2^2 + \lambda
\left|\!\left|\mbox{diag}(s^*)\beta\right|\!\right|_1.\tag{$S$}\label{eq:surrogate_app2}
\end{align}
For a positive vector \(b\), let
\(\bar \beta_{b}(\lambda)\) be a solution to~(\ref{eq:surrogate_app2}) with penalty \(\lambda |\!|\mbox{diag}(b)
\beta|\!|_1\).
\begin{labelpro}{3}[Recoverability of~(\ref{eq:surrogate_app2})]
Suppose \(X\) has absolutely continuous distribution.  For any vectors \(a \geq b \geq {\bf 1}\) and \(\lambda >
0\), w.p. 1
\(\bar\beta_{a}(\lambda), \bar\beta_{b}(\lambda)\) are unique. If additionally \(|\!|\mbox{diag}(a)\bar
\beta_{b}(\lambda)|\!|_1 = |\!|\mbox{diag}(b)\bar \beta_{b}(\lambda)|\!|_1\), then \(\bar
\beta_{a}(\lambda) = \bar \beta_{b}(\lambda)\).
\end{labelpro}
\begin{proof}
Since \(X\) is absolutely continuous, \(a \geq b \geq 0\) and \(\lambda > 0\), it follows that \(\bar\beta_{a}(\lambda),
\bar\beta_{b}(\lambda)\) are almost surely unique~\cite{tibshirani2013lasso}. Since \(a \geq b \geq {\bf 1}\), we
have \(\forall \beta \in \mathbb{R}^p\)
\begin{align}
\frac{1}{2n}\left|\!\left|y - X\beta\right|\!\right|_2^2 + \lambda \left|\!\left|\mbox{diag}(b)\beta\right|\!\right|_1 \leq \frac{1}{2n}\left|\!\left|y - X\beta\right|\!\right|_2^2 + \lambda
\left|\!\left|\mbox{diag}(a)\beta\right|\!\right|_1.
\end{align}
However, we also know
\begin{align}
\lambda
\left|\!\left|\mbox{diag}(b)\bar\beta_{b}(\lambda)\right|\!\right|_1 =
\lambda\left|\!\left|\mbox{diag}(a)\bar\beta_{b}(\lambda)\right|\!\right|_1.
\end{align}
It follows that we must have \(\bar\beta_{a}(\lambda) =
\bar\beta_{b}(\lambda)\).
\end{proof}

\newpage

\section{Proofs of Section 5}
Section 5 compares the estimator of \(\beta^*\) produced by the RepLasso algorithm, with the estimator of
\(\beta^*\) produced by the Lasso. 

The RepLasso is a generalization of the Lasso homotopy method, which maintains a set of weights
\(s(\lambda)\). Indeed, the RepLasso is identical to the Lars algorithm with Lasso modification of Efron et
al.~\cite{efron04lars} if we force \(\theta = {\bf 0}\), which implies that \(\forall \lambda,
s(\lambda) = {\bf 1}\) (We note, however, that for notational convenience the definition of \(\bar w_A\) differs
slightly from that in Efron et al.~\cite{efron04lars} in that case). In the following we will carry out our
comparison of RepLasso with the Lasso homotopy method by comparing the RepLasso with \(\theta \neq {\bf 0}\) and the RepLasso with \(\theta =
{\bf 0}\). We will denote by \(\hat\beta(\lambda)\) the estimator resulting from the
specialization to the Lasso case. Similarly, we let \(\hat w_A\) be the vector corresponding to \(\bar w_A\)
for the Lasso specialization.

We use the following notation inspired by Wainwright~\cite{wainwright09sharp} and Wauthier et
al.~\cite{wauthier2013plasso}. Suppose that the support set of \(\beta^*\) is \(S \triangleq S(\beta^*)\). Let
\(X_j\) be the column \(j\) of \(X\) and \(X_A\) a matrix which consists of the columns indexed by \(A\). For
  all \(j \in S^c\) and \(i \in S\), let
\begin{align}
\mu_j &= X_j^\top X_S (X_S^\top X_S)^{-1} \mbox{sgn}(\beta_S^*)~~~~~~~~\eta_j = X_j^\top \left(I_{n\times n} - X_S (X_S^\top X_S)^{-1} X_S^\top \right) \frac{w}{n}\label{eq:wauthier1}\\
\gamma_i &= e_i^\top \left(\frac{1}{n} X_S^\top X_S\right)^{-1}
\mbox{sgn}(\beta_S^*)~~~~~~~~~\epsilon_i = e_i^\top \left(\frac{1}{n} X_S^\top X_S\right)^{-1} X_S^\top \frac{w}{n}.~~~~~~~~~~~\label{eq:wauthier2}
\end{align}
The proofs of Section 5 use subsets of the following assumptions. 
\begin{itemize}
\itemsep0em
\item[{\bf A1}:] \(\forall G \in \mathcal{G}, |\{i \in G: \beta_i^* \neq 0\}| \leq 1\)
\item[{\bf A2}:] \(\forall A\subset S\) and \(u_A\) the equiangular vector in Eq.~(2.6) of~\cite{efron04lars},  \(\nexists j
  \in A^c, |X_A^\top u_A| = |X_j^\top u_A| {\bf 1}\)
\item[{\bf A3}:] \(X_S^\top X_S\) is invertible
\item[{\bf A4}:] \(|\mu_j| < 1, \forall j \in S^c\), \(\mbox{sgn}(\beta_i^*) \gamma_i > 0~~\forall i \in S\).
\end{itemize}
\newpage
\subsection{Proof of Theorem 2}

\begin{labelthm}{2}[Support Subset Recovery]\label{thm:support_lasso_app}
Assume that {\bf A1--2} hold. Denote by \(\hat\beta(\lambda)\) and \(\bar\beta(\lambda)\) the
Lasso and RepLasso solutions using penalty parameter \(\lambda\). Conditioned on \(X, y\), we have for any \(\lambda_{\min} >
0\)
\begin{align*}
  \forall \lambda \geq \lambda_{\min}~S(\hat\beta(\lambda)) \subseteq S \implies \forall
  \lambda \geq \lambda_{\min}~S(\bar\beta(\lambda)) \subseteq S.
\end{align*}
\end{labelthm}
\begin{proof}
Suppose then that \(\forall \lambda \geq \lambda_{\min}, S(\hat\beta(\lambda)) \subseteq S\). Suppose that
\(\hat \beta(\lambda_{\min})\) corresponds to iteration \(t\) of the Lasso. For \(t' \leq t\), let \(\hat
A_{t'}\) and \(\bar A_{t'}\) be the sequence of active sets of the Lasso and RepLasso up to iteration
\(t\). With a slight abuse of notation we will temporarily treat an active set as an unordered set. Assumption
     {\bf A2} guarantees that for the Lasso, any variable that is at some point in the active set is also at some point in
     the support set. To see this, note that that by {\bf A2}, the vector \(\hat w_A\) never contains a zero
     element. If it did, then an equiangular vector \(u_A\) of \(X_A\) as in Eq.~(2.6) of~\cite{efron04lars}
     could be constructed using a strict subset of vectors indexed by \(A\), violating assumption {\bf
       A2}. But if \(\hat w_A\) does not contain a zero element, then the elements in the active set \(A\)
     cannot indefinitely be assigned a \(\hat\beta_A(\lambda)\) coefficient of zero as \(\lambda\) is swept
     out. Finally, because we know \(\forall \lambda \geq \lambda_{\min}, S(\hat\beta(\lambda)) \subseteq S\),
     this means that \(\forall t' \leq t, \hat A_{t'} \subseteq S\). We will now argue by induction that the
     induced sequence of active sets \(\bar A_{t'}\) of the RepLasso also satisfies \(\forall t' \leq t, \bar
     A_{t'} \subseteq S\).

{\bf Base case:} Since \(s(|\!|X^\top y|\!|_\infty) = {\bf 1}\), the first variable selected by RepLasso and the Lasso
method is the same. That is, \(\hat A_1 = \bar A_1 \subseteq S\) at iteration \(1\).

{\bf Inductive step:} Assume that \(\forall t'' \leq t', \hat A_{t''} = \bar A_{t''} \subseteq S\). Since
\(\forall t'' \leq t', \bar A_{t''} \subseteq S\), we know by {\bf A1} that for all \(\lambda\) up to
iteration \(t'\), \(s(\lambda)\) did not change on \(S\), i.e. \(s_S(\lambda) = {\bf 1}\). This in particular means
that both the Lasso and the RepLasso will have arrived at the same value of \(\lambda\) and intermediate
estimate \(\hat\beta(\lambda) = \bar\beta(\lambda)\) of \(\beta^*\) at the end of stage 4 of iteration
\(t'-1\) and the same vectors \(\hat w_A = \bar w_A\) at the end of stage 2 of iteration \(t'\). To see this,
notice that since \(\forall t'' \leq t', \bar A_{t''} \subseteq S\), and since for all \(\lambda\) up to
iteration \(t'\) we had \(s_S(\lambda) = {\bf 1}\), the RepLasso is up to stage 2 of iteration \(t'\) equivalent to
running Lasso on the subset of variables \(X_{S}, y\).

At stage 3 of iteration \(t'\), the RepLasso algorithm determines whether to add or remove a variable from
\(\bar A_{t'}\) in stage 1 of iteration \(t' + 1\). Since the value of \(\lambda\) and the intermediate
variables \(\hat\beta(\lambda) = \bar\beta(\lambda)\) and \(\hat w_A = \bar w_A\) are the same at stage 2 of
iteration \(t'\) we can now use properties of \(s(\lambda)\) to show that this implies \(\hat A_{t'+1} = \bar
A_{t'+1}\). We consider two cases:
\begin{enumerate}
\item The Lasso determines to add a variable in iteration \(t'+1\) (first bullet in stage 3). Since \(s_S(\lambda) = {\bf 1}\)
  did not change on \(S\), since we always have \(s(\lambda) \geq {\bf 1}\) and since \(\hat A_{t'+1} \subseteq S\),
  it follows that the RepLasso will add the same variable.
\item The Lasso determines to remove a variable in iteration \(t'+1\) (second bullet in stage 3). Since
  \(s_S(\lambda) = {\bf 1}\) did not change on \(S\) and since we always have \(s(\lambda) \geq {\bf 1}\), it then follows
  that the RepLasso will remove the same variable.
\end{enumerate}
Hence, it follows that \(\hat A_{t'+1} = \bar A_{t'+1} \subseteq S\). By the principle of induction, we have
shown that \(\forall t' \leq t, \bar A_{t'} \subseteq S\). 

As the value of \(\lambda\) at the end of stage 4 of iteration \(t\) must be the same for RepLasso and Lasso,
and since for that value we have by definition \(\lambda < \lambda_{\min}\), we now know that \(\forall
\lambda \geq \lambda_{\min}, S(\bar\beta(\lambda)) \subseteq S\).
\end{proof}

\newpage

\subsection{Proof of Theorem 3}
\begin{labelthm}{3}[Signed Support Recovery]
Assume that {\bf A1--4} hold. Denote by \(\hat\beta(\lambda)\) and \(\bar\beta(\lambda)\) the Lasso and RepLasso
solutions using penalty parameter \(\lambda\). For any \(\lambda_{\min} > 0\), we have with probability 1 over
an absolutely continuous distribution on noise \(w\)
\begin{align*}
\forall \lambda \geq \lambda_{\min}~S(\hat\beta(\lambda))
\subseteq S, S_\pm(\hat \beta(\lambda_{\min})) = S_\pm \implies \forall \lambda \geq \lambda_{\min}~S(\bar\beta(\lambda)) \subseteq S, S_\pm(\bar \beta(\lambda_{\min})) = S_\pm.
\end{align*}
\end{labelthm}
\begin{proof}
Suppose then that \(\forall \lambda \geq
\lambda_{\min}~S(\hat\beta(\lambda)) \subseteq S,S_\pm(\hat \beta(\lambda_{\min})) = S_\pm\).  Suppose that
\(\hat\beta(\lambda_{\min})\) corresponds to iteration \(t\) of the Lasso and for \(t' \leq t\) define \(\hat
A_{t'}\) and \(\bar A_{t'}\) to be the active set of Lasso and RepLasso at iteration \(t'\). With a slight abuse
of notation we will temporarily treat an active set as an unordered set. Using the same reasoning as in
Theorem~\ref{thm:support_lasso_app}, it follows from \(\forall \lambda \geq \lambda_{\min}~S(\hat\beta(\lambda)) \subseteq S\)
and {\bf A2} that \(\forall t' \leq t~\hat A_{t'} \subseteq S\) and from this via {\bf A1} that \(\forall t'
\leq t, \bar A_{t'} \subseteq S\). The latter implies that \(\forall \lambda \geq \lambda_{\min},
S(\bar\beta(\lambda)) \subseteq S\). It remains to be shown that \(S_\pm(\bar \beta(\lambda_{\min})) = S_\pm\).

By {\bf A3--4} we assumed that \(X_S^\top X_S\) is invertible, \(|\mu_j| < 1, \forall j \in S^c\) and
\(\mbox{sgn}(\beta_i^*) \gamma_i > 0, \forall i \in S\). We can thus apply Lemma 1 of Wauthier et
al.~\cite{wauthier2013plasso}, which holds with probability 1 over noise instances \(w\). We thus know that
with probability 1, \(S_\pm(\hat \beta(\lambda_{\min})) = S_\pm \iff \lambda_l < \lambda_{\min} < \lambda_u\), where
\begin{align}
 \lambda_l = \max_{j \in S^c} \frac{ \eta_j}{\left(2\llbracket \eta_j >
  0\rrbracket - 1\right) -  \mu_j}~~~~~~~~~~~~~~~~ \lambda_u =
\min_{i \in S} \left|\frac{\beta_i^* +  \epsilon_i}{ \gamma_i}\right|_+, 
\end{align}
\(\llbracket \cdot\rrbracket\) denotes the indicator function and \( \eta_j, \mu_j, \epsilon_i,
\gamma_i\) are defined on \(X, \beta^*, w\) as in Eqs.~(\ref{eq:wauthier1},~\ref{eq:wauthier2}). The RepLasso
algorithm traces out the solution path of a sequence of weighted Lasso problems parameterized by
\(s(\lambda)\)
\begin{align}
\bar \beta(\lambda) &\in \mbox{argmin}_{\beta\in \mathbb{R}^p} \frac{1}{2n}\left|\!\left|y -
X\beta\right|\!\right|_2^2 + \lambda
\left|\!\left|\mbox{diag}(s(\lambda))\beta\right|\!\right|_1.
\end{align}
Suppose we temporarily decouple \(s(\lambda)\)  from the penalty parameter \(\lambda\) and fix it at \(s(\lambda_{\min})\). We
will show that we can then with probability 1 apply Lemma 1 of Wauthier et al.~\cite{wauthier2013plasso} to
the resulting \(s(\lambda_{\min})\)-weighted Lasso problem by applying it to the unweighted Lasso problem on
\(\bar X = X \mbox{diag}(s(\lambda_{\min}))^{-1}, \bar \beta^* = \mbox{diag}(s(\lambda_{\min}))\beta^*, \bar w
= w\). Let \(\bar \eta_j, \bar \mu_j, \bar \epsilon_i, \bar \gamma_i\) be the corresponding variables defined
for \(\bar X, \bar \beta^*, \bar w\). We always have \(s(\lambda_{\min}) \geq {\bf 1}\), and since \(\forall t' \leq
t, \bar A_{t'} \subseteq S\), we also know by {\bf A1} that \(\forall i \in S, s_i(\lambda_{\min}) = 1\). It
follows that \(\bar \eta_j = \eta_j/s_j(\lambda_{\min})\), \(\bar \mu_j = \mu_j / s_j(\lambda_{\min})\), \(\bar \epsilon_i =
\epsilon_i\), \(\bar \gamma_i = \gamma_i\) and \(\bar \beta_i^* = \beta_i^*\). By {\bf A3--4} we assumed that
\(X_S^\top X_S\) is invertible, \(|\mu_j| < 1, \forall j \in S^c\) and \(\mbox{sgn}(\beta_i^*) \gamma_i > 0,
\forall i \in S\). Since \(s(\lambda_{\min}) \geq {\bf 1}\) we have that \(\bar X_S^\top \bar X_S\) is invertible,
\(|\bar \mu_j| < 1, \forall j \in S^c\) and \(\mbox{sgn}(\bar \beta_i^*) \bar \gamma_i > 0, \forall i \in
S\). Hence, we are licensed to apply Lemma~1 of Wauthier et al.~\cite{wauthier2013plasso} to the new problem
instance which produces new bounds \(\bar \lambda_l, \bar\lambda_u\) in terms
of \(\bar \eta_j, \bar \mu_j, \bar \epsilon_i, \bar \gamma_i\). Because \(s(\lambda_{\min}) \geq {\bf 1}\), simple
calculations show that \(\bar \lambda_l \leq \lambda_l\) and \(\lambda_u = \bar
\lambda_u\). Since the Lasso has (with probability 1) \(\lambda_l < \lambda_{\min} <
\lambda_u\), the same \(\lambda_{\min}\) thus also satisfies \(\bar \lambda_l \leq \lambda_l <
\lambda_{\min} < \lambda_u = \bar \lambda_u\). Combining the latter fact with Lemma 1 of
Wauthier et al.~\cite{wauthier2013plasso} we then see that with probability 1 \(S_\pm(\bar
\beta(\lambda_{\min})) = S_\pm\).
\end{proof}

\newpage

\section{RepLars: A RepLasso variant}
As noted in the paper, if \(\theta = {\bf 0}\), and we force \(L = 0\) then the RepLasso algorithm
reduces to the Lars algorithm of Efron et al.~\cite{efron04lars}. (We note, however, that the definition of
\(\bar w_A\) differs slightly from that of the Lars algorithm given in~\cite{efron04lars}). When we force \(L
= 0\) but allow \(\theta \neq {\bf 0}\) we have a new algorithm, which we call RepLars. In this section we present
this algorithm and analyze its behavior. 

\begin{algorithm}[h]
{
\begin{pseudocode}{REPLARS}{\(X, y, \mathcal{G}, \theta\)}\label{alg:replars}
\bar y = 0, A = (), \lambda = |\!|X^\top y|\!|_\infty, s(\lambda) = {\bf 1}, \bar\beta(\lambda) =
0\todo{change \(s(\lambda)\) to \(\bar s(\lambda)?, \(A_A\) to \(\bar A_A\), etc}\\
\WHILE \lambda > 0\DO 
\BEGIN
\mbox{Stage 1}\left\{\BEGIN
A = (A, i^*),~\mbox{where}~i^* = \mbox{argmax}_{j \in A^c} \left|X_j^\top (y - \bar y)/s_j(\lambda)\right|\\

s_M(\lambda^-) = s_M(\lambda) + \frac{\theta_{\Gamma(i^*)}}{|G_{\Gamma(i^*)}| - 1}{\bf 1},~~~s_{M^c}(\lambda^-) = s_{M^c}(\lambda),~~\mbox{with}~~M = \left\{A^c \cap G_{\Gamma(i^*)}\right\}\\
\END\right.\\
\mbox{Stage 2}\left\{\BEGIN
\bar w_A =  A_A \left( X_A^\top X_A \right)^{-1} \mbox{diag}(\mbox{sgn}\left(X_A^\top (y - \bar y)\right))
s_A(\lambda),~\mbox{with}~A_A~\mbox{s.t.}~|\!|X_A \bar w_A|\!|_2^2 = 1\\
\END\right.\\
\mbox{Stage 3}\left\{\BEGIN
\mbox{Find smallest \(\rho > 0\) s.t.}~\exists j \in A^c~\mbox{s.t.}~|X_j^\top (y - \bar y - \rho X_A \bar w_A) / s_j(\lambda)| = \lambda -
\rho\!\\

\END\right.\\
\mbox{Stage 4}\left\{\BEGIN
\bar\beta_A(\lambda - \rho) = \bar\beta_A(\lambda) + \rho \bar w_A,~~~\bar\beta_{A^c}(\lambda - \rho) = 0,~~~\bar y = X \bar\beta(\lambda - \rho) \\
\lambda = \lambda - \rho\\
\END\right.\\
\END\\
\RETURN \bar\beta
\end{pseudocode}
}
\end{algorithm}
In the following we will compare of RepLars with the Lars by comparing RepLars with \(\theta > 0\) and RepLars with
\(\theta = {\bf 0}\). We will denote by \(\hat\beta\) the Lars estimate corresponding to \(\bar \beta\)
produced by the algorithm above. Similarly, we let \(\hat w_A\) be the vector in the Lars specialization
corresponding to \(\bar w_A\) above. We will assume throughout this analysis that variables are added to the
active set one by one\todo{Requires that order statistics of \(y\) have no ties?}. 

\newpage
\begin{labelthm}{4}
Assume that {\bf A1--2} hold. Denote by \(\hat \beta(\lambda)\) and \(\bar \beta(\lambda)\) the Lars and
RepLars solutions indexed by parameter \(\lambda\). Conditioned on \(X, y\), we have for any \(\lambda_{\min}
> 0\)
  \begin{align}
    \forall \lambda \geq \lambda_{\min}~S(\hat \beta(\lambda)) \subseteq S \implies \forall \lambda \geq \lambda_{\min}~S(\bar \beta(\lambda)) \subseteq S.
  \end{align}
\end{labelthm}
\begin{proof}
  Suppose that \(\forall \lambda \geq \lambda_{\min}~S(\hat \beta(\lambda)) \subseteq S\). Suppose further
  that \(\hat\beta(\lambda_{\min})\) corresponds to iteration \(t\) of Lars. For \(t' \leq t\) let \(\hat
  A_{t'}\) be the active sets of Lars at iteration \(t'\). With a slight abuse of notation we will temporarily
  treat an active set as an unordered set. Assumption {\bf A2} guarantees that any variable that is at some
  point in the active set is also at some point in the support set. To see this, note that that by {\bf A2},
  the vector \(\hat w_A\) never contains a zero element. If it did, then an equiangular vector \(u_A\) of
  \(X_A\) as in Eq.~(2.6) of~\cite{efron04lars} could be constructed using a strict subset of vectors indexed
  by \(A\), violating assumption {\bf A2}. But if \(\hat w_A\) does not contain a zero element, then the
  elements in the active set \(A\) cannot indefinitely be assigned a \(\hat\beta_A(\lambda)\) coefficient of
  zero as \(\lambda\) is swept out. Finally, because we know \(\forall \lambda \geq \lambda_{\min},
  S(\hat\beta(\lambda)) \subseteq S\), this means that \(\forall t' \leq t, \hat A_{t'} \subseteq S\). We will
  now argue by induction that the induced sequence of active sets \(\bar A_{t'}\) of the RepLars also
  satisfies \(\forall t' \leq t, \bar A_{t'} \subseteq S\).

{\bf Base case:} Since \(s(|\!|X^\top y|\!|_\infty) = {\bf 1}\), the first variable selected by RepLars and the Lars
method is the same. That is, \(\hat A_1 = \bar A_1 \subseteq S\) at iteration \(1\).

{\bf Inductive step:} Assume that \(\forall t'' \leq t', \hat A_{t''} = \bar A_{t''} \subseteq S\). Since
\(\forall t'' \leq t', \bar A_{t''} \subseteq S\), we know by {\bf A1} that for all \(\lambda\) up to
iteration \(t'\), \(s(\lambda)\) did not change on \(S\), i.e. \(s_S(\lambda) = {\bf 1}\). This in particular means
that both the Lars and the RepLars will have arrived at the same value of \(\lambda\) and intermediate
estimate \(\hat\beta(\lambda) = \bar\beta(\lambda)\) of \(\beta^*\) at the end of stage 4 of iteration
\(t'-1\) and the same vectors \(\hat w_A = \bar w_A\) at the end of stage 2 of iteration \(t'\). To see this,
notice that since \(\forall t'' \leq t', \bar A_{t''} \subseteq S\), and since for all \(\lambda\) up to
iteration \(t'\) we had \(s_S(\lambda) = {\bf 1}\), the RepLars is up to stage 2 of iteration \(t'\) equivalent to
running Lars on the subset of variables \(X_{S}, y\).

At stage 3 of iteration \(t'\), the RepLars algorithm determines which variable to add to \(\bar A_{t'}\) in
stage 1 of iteration \(t' + 1\). Since the value of \(\lambda\) and the intermediate variables
\(\hat\beta(\lambda) = \bar\beta(\lambda)\) and \(\hat w_A = \bar w_A\) are the same at stage 2 of iteration
\(t'\) we can now use properties of \(s(\lambda)\) to show that this implies \(\hat A_{t'+1} = \bar
A_{t'+1}\). Specifically, since (1) \(s_S(\lambda) = {\bf 1}\) did not change on \(S\); (2) we always have
\(s(\lambda) \geq {\bf 1}\); and (3) \(\hat A_{t'+1} \subseteq S\), it follows that the RepLars will add the same
variable. Hence, it follows that \(\hat A_{t'+1} = \bar A_{t'+1} \subseteq S\).

By the principle of induction, we have shown that \(\forall t' \leq t, \bar A_{t'} \subseteq S\). As the value
of \(\lambda\) at the end of stage 4 of iteration \(t\) must be the same for RepLars and Lars, and since for
that value we have by definition \(\lambda < \lambda_{\min}\), we now know that \(\forall \lambda \geq
\lambda_{\min}, S(\bar\beta(\lambda)) \subseteq S\).
\end{proof}